\documentclass[10pt,twocolumn,letterpaper]{article}

\usepackage{cvpr}              % To produce the CAMERA-READY version
\usepackage{amsmath}
\usepackage{multirow}
\usepackage{subcaption}
\usepackage{graphicx}
\usepackage{pifont}
\usepackage[ruled,vlined,linesnumbered]{algorithm2e}

\usepackage[utf8]{inputenc} % allow utf-8 input
\usepackage[T1]{fontenc}    % use 8-bit T1 fonts
\usepackage{url}            % simple URL typesetting
\usepackage{booktabs}       % professional-quality tables
\usepackage{amsfonts}       % blackboard math symbols
\usepackage{nicefrac}       % compact symbols for 1/2, etc.
\usepackage{microtype}      % microtypography
\usepackage{xcolor}         % colors

\definecolor{commentGreen}{rgb}{0,0.5,0.05}
\newcommand{\xmark}{\ding{55}}

\definecolor{cvprblue}{rgb}{0.21,0.49,0.74}
\usepackage[pagebackref,breaklinks,colorlinks,allcolors=cvprblue]{hyperref}

\def\paperID{10099} % *** Enter the Paper ID here
\def\confName{CVPR}
\def\confYear{2025}

\title{Curriculum Direct Preference Optimization for Diffusion and Consistency Models\vspace{-0.3cm}}

\author{Florinel-Alin Croitoru$^{1}$, Vlad Hondru$^1$, {Radu Tudor} Ionescu$^{1,}$\thanks{Corresponding author: \texttt{raducu.ionescu@gmail.com}.} , Nicu Sebe$^2$, Mubarak Shah$^3$\\
$^1$University of Bucharest, Romania, $^2$University of Trento, Italy,
$^3$University of Central Florida, US\vspace{-0.3cm}
}

\begin{document}

\maketitle

\begin{abstract}
Direct Preference Optimization (DPO) has been proposed as an effective and efficient alternative to reinforcement learning from human feedback (RLHF). In this paper, we propose a novel and enhanced version of DPO based on curriculum learning for text-to-image generation. Our method is divided into two training stages. First, a ranking of the  examples generated for each prompt is obtained by employing a reward model. Then, increasingly difficult pairs of examples are sampled and provided to a text-to-image generative (diffusion or consistency) model. Generated samples that are far apart in the ranking are considered to form easy pairs, while those that are close in the ranking form hard pairs. In other words, we use the rank difference between samples as a measure of difficulty. The sampled pairs are split into batches according to their difficulty levels, which are gradually used to train the generative model. Our approach, Curriculum DPO, is compared against state-of-the-art fine-tuning approaches on nine benchmarks, outperforming the competing methods in terms of text alignment, aesthetics and human preference. Our code is available at \url{https://github.com/CroitoruAlin/Curriculum-DPO}.
\end{abstract}

\setlength{\abovedisplayskip}{2.0pt}
\setlength{\belowdisplayskip}{2.0pt}

\vspace{-0.2cm}
\section{Introduction}
\label{sec: introduction}
\vspace{-0.1cm}

Diffusion models \cite{Croitoru-TPAMI-2023,ho-NeurIPS-2020,sohl-icml-2015,song-NeurIPS-2019} represent a family of generative models that gained significant traction in image generation tasks, largely due to their impressive generative capabilities. One of the tasks where these models excel is text-to-image generation \cite{avrahami-CVPR-2022,gu-CVPR-2022,rombach-CVPR-2022,Saharia-NeurIPS-2022}, as they are capable of generating images that are both aesthetic and well aligned with the input text (prompt). However, state-of-the-art diffusion models that have been widely adopted by the community, e.g.~Stable Diffusion \cite{rombach-CVPR-2022}, are typically heavy in terms of training (and even inference) time. To this end, several studies \cite{Black-ICLR-2024,Fan-NeurIPS-2023,Luo-Arxiv-2023b,Wallace-arxiv-2023} proposed novel training methods to efficiently fine-tune pre-trained diffusion models. Some of these training frameworks \cite{Black-ICLR-2024,Luo-Arxiv-2023b,Wallace-arxiv-2023} were originally introduced for Large Language Models (LLMs) \cite{Christiano-NeurIPS-2017,Hu-ICLR-2022,Rafailov-NeurIPS-2023}, another family of models that are notoriously hard to train on a few GPUs \cite{Schwartz-CACM-2020,Strubell-ACL-2019}. This is also the case of Direct Preference Optimization (DPO) \cite{Rafailov-NeurIPS-2023}, a method used to fine-tune LLMs, which was originally proposed as an effective and efficient alternative to reinforcement learning from human feedback (RLHF) \cite{Christiano-NeurIPS-2017}. DPO bypasses the need to fit a reward model by harnessing a mapping between reward functions and optimal policies, which enables the direct optimization of the LLM to adhere to human preferences. DPO was later extended to diffusion models \cite{Wallace-arxiv-2023}, showcasing similar benefits in image generation. Despite the significant benefits brought by Diffusion-DPO \cite{Wallace-arxiv-2023} and related methods \cite{Black-ICLR-2024,Fan-NeurIPS-2023,Luo-Arxiv-2023b}, there are still observable gaps in terms of various factors, such as text alignment, aesthetics and human preference (see Figures \ref{qualitative_text_align},  \ref{qualitative_human_preference} and \ref{qualitative_aesthetics}). Therefore, more advanced adaptation techniques are required to reduce these gaps. 

\begin{figure*}[t]
  \centering
  \includegraphics[width=0.825\textwidth]{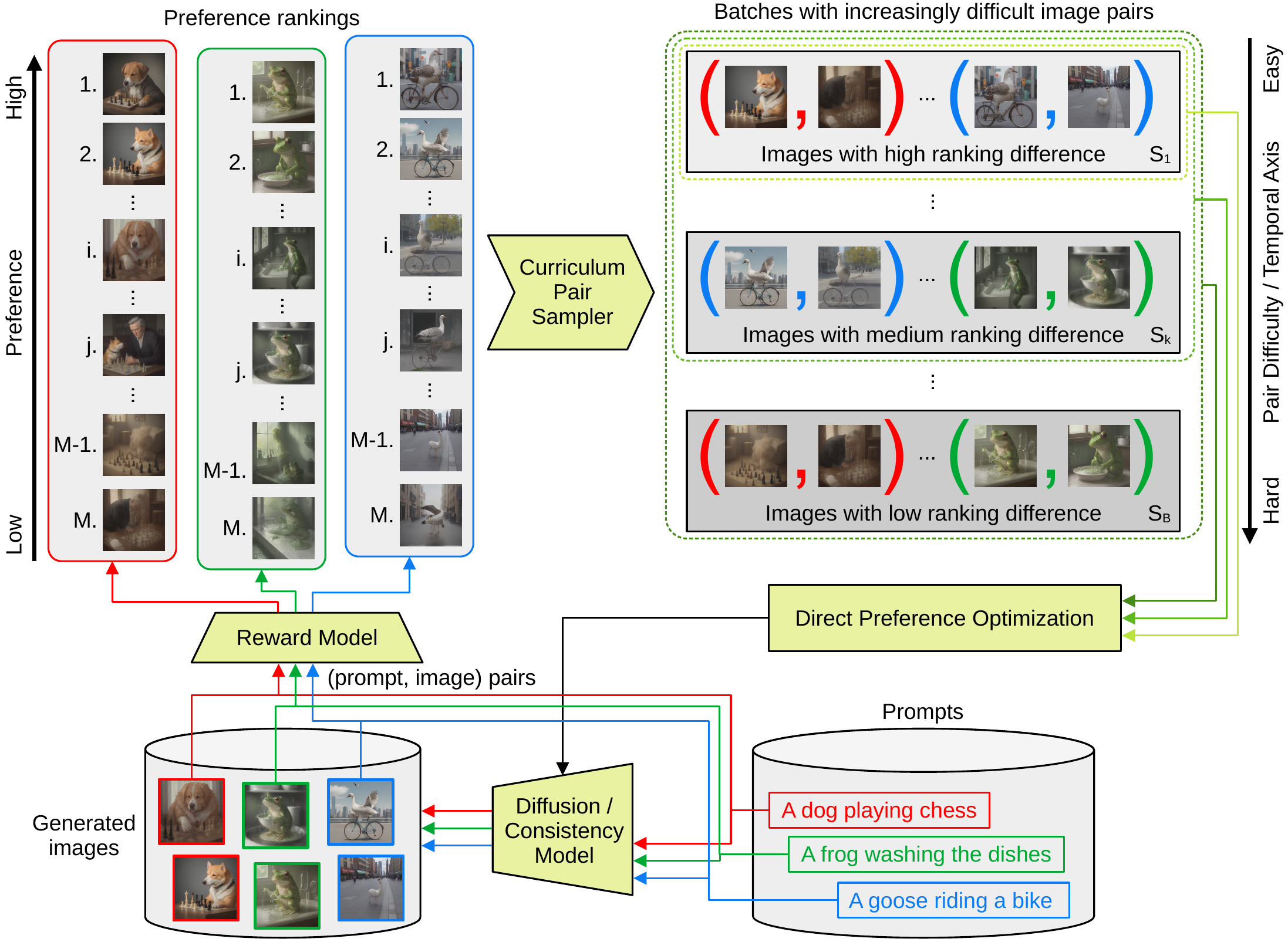}
  %\fbox{\rule[-.5cm]{0cm}{4cm} \rule[-.5cm]{4cm}{0cm}}
  \vspace{-0.25cm}
  \caption{An overview of Curriculum DPO. Images generated by a diffusion / consistency model and their prompts are passed through a reward model, obtaining a preference ranking for each prompt. Next, image pairs of various difficulty levels are generated and organized into batches, such that the initial batch contains easy pairs (with high difference in terms of preference scores) and subsequent batches contain increasingly difficult pairs (the difference in terms of preference is gradually decreased). The diffusion / consistency model is finally trained via Direct Preference Optimization (DPO) based on curriculum learning. Best viewed in color.\vspace{-0.4cm}}
  \label{fig_pipeline}
\end{figure*}

%\vspace{-0.1cm}
In this work, we propose a novel and enhanced version of Direct Preference Optimization based on curriculum learning \cite{Bengio-ICML-2009,Soviany-IJCV-2022}, which is termed \textbf{Curriculum DPO}, to train text-to-image generative (diffusion \cite{rombach-CVPR-2022} or consistency \cite{Song-ICLR-2024,Song-ICML-2023}) models. Using Diffusion-DPO \cite{Wallace-arxiv-2023}, the model is trained to learn the preferred example from a pair of generated examples. However, the pairs are randomly sampled during training, which is suboptimal. We conjecture that organizing the pairs according to their complexity with respect to the preference learning task, from easy to hard, can lead to a more effective training process. Our training process comprises two stages. In the first stage, we employ a reward model to rank the images generated for each prompt according to their preference score. In the second stage, we create pairs of samples with different difficulty levels, leveraging the rank difference between samples as a measure of difficulty. Samples that are far apart in the preference ranking are considered to form easy pairs, as we naturally assume that it is more obvious which is the preferred sample. In contrast, samples that are close in the ranking form hard pairs, since choosing the preferred sample is less obvious. The sampled pairs are split into batches according to their difficulty level, which allows us to organize the batches in a meaningful order, from easy to hard. This essentially generates an optimization process based on curriculum learning \cite{Bengio-ICML-2009}, as illustrated in Figure \ref{fig_pipeline}. Our study is motivated by the success of various curriculum learning methods employed in different domains \cite{Jimenez-MICCAI-2019,Liu-IJCAI-2018,Madan-WACV-2024,Sinha-NIPS-2020,Soviany-CVIU-2021,Wei-WACV-2021}. However, to the best of our knowledge, our work is the first to employ a curriculum learning strategy to fine-tune (adapt) diffusion and consistency models for text-to-image generation. 

%\vspace{-0.1cm}
We carry out experiments across nine evaluation benchmarks to assess the effectiveness of Curriculum DPO in adapting text-to-image generative models to different reward models. We first tackle the text alignment task, employing the Sentence-BERT \cite{reimers-EMNLP-2019} similarity metric to evaluate the correspondence between the actual text prompt and the caption provided by the LLaVA model \cite{liu-NeurIPS-2024} for each generated image. Next, we focus on enhancing the visual aesthetics of the images, where we use the LAION Aesthetics Predictor \cite{Schuhmann-laion-2022} as the reward model. Finally, for the third task, we increase the human preference rate by employing HPSv2 \cite{Wu-arXiv-2023} as the reward model. Each of the three tasks is studied on two distinct datasets. The quantitative and qualitative results on all datasets show that Curriculum DPO surpasses state-of-the-art fine-tuning strategies, such as Diffusion-DPO \cite{Rafailov-NeurIPS-2023, Wallace-arxiv-2023} and Denoising Diffusion Policy Optimization (DDPO) \cite{Black-ICLR-2024}. %In terms of quantitative metrics, Curriculum DPO showed higher scores in alignment with the Sentence-BERT for text relevance, improved ratings via the LAION Aesthetics Predictor for visual appeal, and stronger alignment with human preferences according to HPSv2.
Furthermore, we conduct a study based on actual human feedback to further evaluate the fine-tuning methods. %trained with DPO \cite{Rafailov-NeurIPS-2023, Wallace-arxiv-2023}, Curriculum DPO, and DDPO\cite{Black-ICLR-2024}.
%A group of evaluators rated the images produced by each method according to two criteria: text alignment and visual appeal.
The outcome of this study reinforces the findings from our quantitative analysis, showing that Curriculum DPO consistently generates the most preferred samples out of the three methods. Moreover, our ablation study on the number of training samples shows that Curriculum DPO achieves similar performance to Diffusion-DPO and DDPO, while using 10$\times$ less images. Overall, our results demonstrate the effectiveness of Curriculum DPO. By introducing increasingly complex pairs, Curriculum DPO ensures that the generative models progressively refine their capabilities, leading to outputs that are better aligned with the reward models.

%\vspace{-0.1cm}
In summary, our contribution is threefold:
\begin{itemize}
    \item %\vspace{-0.1cm}
    We introduce Curriculum DPO, a novel training regime for diffusion and consistency models that enhances DPO via curriculum learning.
    \item %\vspace{-0.1cm}
    We propose an adaptation of DPO for consistency models, termed Consistency-DPO, enabling short training and inference times.
    \item %\vspace{-0.1cm}
    We demonstrate the superiority of Curriculum DPO over state-of-the-art training alternatives on nine evaluation benchmarks.
\end{itemize}

\vspace{-0.1cm}
\section{Related Work}
\label{sec: related_work}

\vspace{-0.1cm}
\textbf{Diffusion Models.}
A new class of generative models, known as diffusion models, has emerged over the last few years \cite{Croitoru-TPAMI-2023, ho-NeurIPS-2020, rombach-CVPR-2022, sohl-icml-2015, song-NeurIPS-2019, song-ICLR-2021}, rapidly gaining traction due to their capability in synthesizing high-quality and diverse images. This has lead to many ongoing research directions, ranging from the usage of various modalities for guidance (\eg~text \cite{avrahami-CVPR-2022,gu-CVPR-2022,ramesh-arXiv-2022,rombach-arXiv-2022}, image \cite{baranchuk-arXiv-2021, meng-arXiv-2021, saharia-SIGGRAPH-2022} or specific class \cite{chao-ICLR-2022, ho-arXiv-2021, salimans-arXiv-2022}) to the application in many visual tasks, such as inpainting \cite{lugmayr-CVPR-2022, nichol-ICML-2022} and super-resolution \cite{daniels-NeurIPS-2021, saharia-TPAMI-2022}, as well as different domains, such as medical imaging \cite{chung-MIA-2022, wyatt-CVPR-2022}. The main drawback of diffusion models is the sampling time, because they typically require a large number of denoising iterations. Significant effort has been dedicated to optimizing the scheduler \cite{liu-ICLR-2022, nichol-ICML-2021, song-ICLR-2021b}. This effort has led to a new branch of diffusion models, called consistency models, being introduced by \citet{Song-ICML-2023}. Consistency models were further improved in several recent works, \eg~\cite{Luo-arXiv-2023, Song-ICLR-2024}. To the best of our knowledge, DPO was not studied in conjunction with consistency models.

% \vspace{-0.1cm}
\noindent
\textbf{Controllable generation for diffusion models.}
Substantial endeavors focused on guiding the diffusion process in order to better control the synthesized images. Early attempts used the gradients of a classifier \cite{chao-ICLR-2022, dhariwal-NeurIPS-2021, song-ICLR-2021}, after which \citet{ho-NeurIPS-2021} proposed a classifier-free guidance strategy for training conditional diffusion models. 
\citet{Zhang-ICCV-2023} presented ControlNet, a method that adds a conditioning input to any text-to-image diffusion model. Their method involves architectural modifications: while keeping the original U-Net intact and frozen, trainable copies of the encoder blocks and middle blocks are created for the control input, which are then integrated in the original decoder blocks with zero-convolutional layers.

% \vspace{-0.1cm}
Recent breakthroughs in LLMs were adopted and applied to diffusion models. \citet{Hu-ICLR-2022} proposed an efficient method to fine-tune LLMs by introducing a trainable low-rank matrix decomposition for certain parameters of transformer layers, and then summing them, while keeping the original weights frozen. \citet{Luo-Arxiv-2023b} employed Low-Rank Adaptation (LoRA) \cite{Hu-ICLR-2022} on consistency models, demonstrating better results while requiring less memory.
The loss function for training LLMs is not capturing human preference, while current metrics (\eg~BLEU \cite{papineni-ACL-2002}) defectively assess it. To this end,  reinforcement learning methods have been adopted \cite{bai-arXiv-2022, havrilla-arXiv-2024, Ziegler-arXiv-2020}. By learning a reward function from a dataset collected with human feedback and fine-tuning a text-to-image diffusion model on it, \citet{lee-ArXiv-2023} showed that the generated images are better aligned with the input prompt. Fine-tuning LLMs using policy gradient algorithms has played a major role in their success, and thus, they were rapidly adopted in diffusion models as well. While \citet{Fan-NeurIPS-2023} demonstrated that such an algorithm leads to an improved text-image alignment, \citet{Black-ICLR-2024} applied their method on enhancing multiple aspects of synthesis: compressibility, aesthetics, as well as prompt alignment. In their recent work, \citet{Rafailov-NeurIPS-2023} presented Direct Preference Optimization, an algorithm for fine-tuning LLMs from any preference, proving to be superior to Proximal Policy Optimization methods \cite{ouyang-NeurIPS-2022, schulman-arXiv-2017, Ziegler-arXiv-2020}, and thus, replacing RLHF algorithms. It involves directly training the model by modifying the objective to integrate the preference rather than first fitting a reward model and then using it to train the original model. \citet{Wallace-arxiv-2023} reformulated DPO to fit the context of diffusion models by employing a novel training strategy and loss function. Their experiments demonstrate promising results on subjective metrics such as visual aesthetics or prompt alignment. Nevertheless, to the best of our knowledge, curriculum learning has not been applied in conjunction with controllable generation methods for diffusion models, such as DPO \cite{Wallace-arxiv-2023} and DDPO \cite{Black-ICLR-2024}.

%\vspace{-0.1cm}
\noindent
\textbf{Curriculum learning.} The process of training neural networks on samples with increasing difficulty at each iteration is known as \emph{curriculum learning}. Although curriculum learning was introduced almost 15 years ago by \citet{Bengio-ICML-2009}, it is still actively integrated in many recent methods with great success \cite{Soviany-IJCV-2022}. 
There are various strategies for implementing curriculum learning: either progressively inserting harder samples in the training data \cite{Bengio-ICML-2009, jiang-AAAI-2015, kuman-NeurIPS-2010} or making the training objective more difficult \cite{karras-ICLR-2018, morerio-ICCV-2017}. While curriculum learning has been extensively applied in computer vision \cite{Soviany-CVIU-2021, Wang-ICCV-2023}, it was less utilized in image generation, mostly on Generative Adversarial Networks \cite{doan-AAAI-2019, ghasedi-CVPR-2019, karras-ICLR-2018, soviany-wacv-2020}. The easy-to-hard learning strategy was only recently incorporated in diffusion models \cite{kim-arXiv-2024,xu-arXiv-2024}. \citet{kim-arXiv-2024} and \citet{xu-arXiv-2024} advocate starting the training from higher timesteps with more noise and progressing to lower timesteps with less noise. Their methods are completely different from our work, as we concentrate on the difficulty with respect to the preference scores rather than the noise levels.

\vspace{-0.1cm}
\section{Method}
\label{sec: method}

% \textbf{Overview}
\vspace{-0.1cm}
We begin by introducing the mathematical preliminaries behind diffusion models, consistency models, DPO and Diffusion-DPO (for a more detailed introduction of these concepts, please refer to Appendix \ref{supp_prelim}). We then present our adaptation of DPO for consistency models. Finally, we delve into our curriculum learning strategy designed for DPO.

% In this section, we begin by discussing the diffusion and consistency models. We then delve into the methodologies of DPO and Diffusion-DPO. Finally, we conclude the section with a presentation of our curriculum learning strategy designed for DPO and our adaptation of DPO for consistency models.
\vspace{-0.1cm}
\subsection{Preliminaries}
\vspace{-0.1cm}
\textbf{Diffusion models.} Diffusion models are a category of generative models that learn to reverse a process, called the forward process, where Gaussian noise is added to the original samples $x_0 \sim p(x_0)$ over $T$ steps, guided by a predefined schedule defined by $(\alpha_t)_{t=1}^T$ and $(\sigma_{t=1}^T)$: $x_t = \alpha_t x_0 + \sigma_t \epsilon$, where $\epsilon \sim \mathcal{N}(0, \mathbf{I})$. These models are conditioned on the time step $t$ and trained to estimate the inserted noise $\epsilon$ through a denoising objective:
\begin{equation}
    \!\!\mathcal{L}_{\mbox{\scriptsize{simple}}}\!=\! \mathbb{E}_{t \sim \mathcal{U}(1, T), \epsilon \sim \mathcal{N}(0, \mathbf{I}), x_0 \sim p(x_0)} \!\left\lVert \epsilon_t\!-\!\epsilon_\theta(x_t,t)\right\rVert^{2}_2\!,
\end{equation}
where $\epsilon_\theta(x_t, t)$ represents the time-dependent neural network and $\theta$ are the trainable parameters.
The generation of new samples with diffusion models is a multi-step process where an initial Gaussian sample is progressively transformed into one from the original distribution. At each step, the model $\epsilon_\theta$ estimates and subtracts the noise, refining the sample towards the target with each iteration.

%\vspace{-0.1cm}
\noindent
\textbf{Consistency models.} %\cite{Luo-arXiv-2023, Song-ICLR-2024,  Song-ICML-2023}
\citet{song-ICLR-2021} showed that the usual stochastic denoising process of diffusion models has an equivalent deterministic process, described by an ordinary differential equation (ODE), called Probability flow-ODE (PF-ODE).
\citet{Song-ICML-2023} introduced consistency models \cite{Luo-arXiv-2023, Song-ICLR-2024, Song-ICML-2023} based on the idea of training a neural network to map each point along a solution trajectory of the PF-ODE to its initial point, representing the original sample. To this end, \citet{Song-ICML-2023} proposed a training objective which enforces the model, $f_\phi(x_t, t)$, to have the same output, given two points of the trajectory:
\begin{equation}
\label{eq_consistency_distillation}
    \mathcal{L}_{\mbox{\scriptsize{CD}}}(\phi) = d(f_\phi(x_{t_{n+1}}, t_{n+1}), f_{\phi^{-}}(\hat{x}_{t_n}^{\theta}, t_n)),
\end{equation}
where $d$ is a distance metric, $n \sim \mathcal{U}(1, N)$, $N$ is the discretization length of the interval $\left[0, T\right]$, $\phi$ are the trainable parameters of the consistency model and $\phi^{-}$ is a running average of $\phi$. The term $\hat{x}_{t_n}^\theta$ represents a one-step denoised version of $x_{t_{n+1}}$, obtained by applying an ODE solver on the PF-ODE. The solver operates using a pre-trained diffusion model, $\epsilon_\theta(x_{t_n}, t_n)$.

% \vspace{-0.1cm}
\noindent
\textbf{Direct Preference Optimization (DPO).} 
\citet{Rafailov-NeurIPS-2023} introduced DPO, a method to fine-tune LLMs with pairs of ranked examples $(x_0^w, x_0^l, c)$, where $x_0^w$ is preferred, $x_0^l$ is less favored, and $c$ is a condition under which both samples are generated. The training objective aims to increase the likelihood of the preferred examples, $p_\theta(x_0^w|c)$, and decrease that of less preferred ones, while not diverging from the initial state of the model, denoted by $p_{\mbox{\scriptsize{ref}}}(x_0|c)$:
\begin{equation}
    \label{eq_dpo}
    \begin{split}
        \mathcal{L}_\mathrm{\mbox{\scriptsize{DPO}}}(\theta)\!\!=\!\!-\!\mathbb{E}_{x_0^w\!, x_0^l, c} \!\left[\!\log\!\sigma\!\bigg(\!\!\beta\!\! %\cdot
        \left(\!\!\log\!{\frac{p_\theta(x_0^w\!|c)}{p_{\mbox{\scriptsize{ref}}}(x_0^w\!|c)}}\!-\!\log\!{\frac{p_\theta(x_0^l\!|c)}{p_{\mbox{\scriptsize{ref}}}(x_0^l\!|c)}}\!\!\right)\!\!\!\!\bigg)\!\!\right]\!\!,
    \end{split} 
\end{equation}
where $\sigma$ is the sigmoid function and $\beta$ is a hyperparameter controlling the divergence of $p_\theta(x_0|c)$ from $p_{\mbox{\scriptsize{ref}}}(x_0|c)$. To better grasp the intuition behind $\mathcal{L}_\mathrm{\mbox{\scriptsize{DPO}}}$, we can look at its gradient with respect to $\theta$:
\begin{equation}
    \label{eq_grad_dpo}
    \begin{split}
        \frac{\partial \mathcal{L}_\mathrm{\mbox{\scriptsize{DPO}}}(\theta)}{\partial\theta}\!&=\! -\beta \mathbb{E}_{x_0^w, x_0^l, c}\Bigg[ \sigma\left( \hat{r}_\theta(x_0^l, c) - \hat{r}_\theta(x_0^w, c)\right)\cdot\\ &\left( \frac{\partial\log{p_\theta(x_0^w|c)}}{\partial\theta} - \frac{\partial\log{p_\theta(x_0^l|c)}}{\partial\theta} \right)\!\Bigg],
    \end{split}
\end{equation}
where $\hat{r}_{\theta}(x_0,c)=\beta\cdot\log{\frac{p_\theta(x_0|c)}{p_{\mbox{\tiny{ref}}}(x_0|c)}}$. By analyzing Eq.~\eqref{eq_grad_dpo}, we can observe that the DPO objective enhances the likelihood of preferred examples while diminishing it for the unfavored ones. Moreover, the update is weighted by a sigmoid term that is proportional to how wrong the model is.

%\vspace{-0.1cm}
\noindent
\textbf{Diffusion-DPO}. \citet{Wallace-arxiv-2023} applied DPO to diffusion models by modifying Eq.~\eqref{eq_dpo}. They replaced the logarithmic differences, approximating them with differences between the errors on noise estimates:
\begin{equation}
    \label{eq_diffusion_dpo}
    \begin{split}
        \!\mathcal{L}_{\mathrm{\mbox{\scriptsize{Diff-DPO}}}}&(\theta) \!=\! - \mathbb{E}_{x_t^w, x_t^l, c}\!\Big[\!\log\sigma\Big(\!\!-\!\beta \cdot T %\omega(\lambda_t)
        \Big(\\ &
    %\left(
    \lVert\epsilon^w\!-\!\epsilon_\theta^w(x_t^w\!, t, c)\rVert_2^2\!-\! 
    \lVert\epsilon^w\!-\!\epsilon_{\mbox{\scriptsize{ref}}}^w(x_t^w, t, c)\rVert_2^2 -\\ &
    %\right) - \\ &
    %\left(
    \lVert\epsilon^l\!-\!\epsilon_\theta^l(x_t^l, t, c)\rVert_2^2\!+\! 
    \lVert\epsilon^l\!-\!\epsilon_{\mbox{\scriptsize{ref}}}^l(x_t^l, t, c)\rVert_2^2
    % \right)
    \!\Big)\!\Big)\!\Big]\!,
    \end{split}
\end{equation}
where $x_t^w$ and $x_t^l$ are obtained  from $x_0^w$ and $x_0^l$ using the forward process of diffusion models.%Eq.~\eqref{eq_forward_process}. %$\omega(\lambda_t)$ is a weighting function dependent on the signal-to-noise ratio $\lambda_t = \frac{\alpha_t^2}{\sigma_t^2}$.

\vspace{-0.1cm}
\subsection{Consistency-DPO} 
\vspace{-0.1cm}
Inspired by the DPO implementation of \citet{Wallace-arxiv-2023} for diffusion models, we propose a parallel adaptation for consistency models. Specifically, we utilize Eq.~\eqref{eq_consistency_distillation} to measure and enhance the model's accuracy on preferred examples, while permitting a rise in error rates for less favored ones. Given a reference pre-trained consistency model $f_{\mbox{\scriptsize{ref}}}$, we propose the following DPO objective to fine-tune on ranked examples:
\begin{equation}
    \label{eq_consistency_dpo}
    \begin{split}
        \mathcal{L}_\mathrm{\mbox{\scriptsize{Con-DPO}}}&(\phi)\!=\! -\mathbb{E}_{x_{t_{n+1}}^w, x_{t_{n+1}}^l, c} \Big[\log \sigma\Big(\!-\beta\Big(\\ & d^w(x_{t_{n+1}}^w,\hat{x}_{t_n}^{w,\theta}, \phi) - d^l(x_{t_{n+1}}^l,\hat{x}_{t_n}^{l,\theta}, \phi)\!\Big)\!\Big)\!\Big], 
    \end{split}
\end{equation}
where, for $* \in \{w, l\}$, $d^*$ is defined as:
\begin{equation}
    \label{eq_d_star}
    \begin{split}
       d^* = &\,d(f_\phi(x_{t_{n+1}}^*, t_{n+1}, c), f_{\mbox{\scriptsize{ref}}}(\hat{x}_{t_{n}}^{*,\theta}, t_{n}, c)) - \\ & \,d(f_{\mbox{\scriptsize{ref}}}(x_{t_{n+1}}^*, t_{n+1}, c), f_{\mbox{\scriptsize{ref}}}(\hat{x}_{t_{n}}^{*,\theta}, t_{n}, c)). 
    \end{split}
        % -\\
        % &\left(d(f_\phi(x_{t_{n+1}}^l, t_{n+1}, c), f_{ref}(\hat{x}_{t_{n}}^{l,\theta}, t_{n}, c)) - d(f_{ref}(x_{t_{n+1}}^l, t_{n+1}, c), f_{ref}(\hat{x}_{t_{n}}^{l,\theta}, t_{n}, c))\right) \\
        % &\Big)\Big) \Big]
\end{equation}
Note that $c$ is a condition, as in previous cases. The variable $x_{t_{n+1}}^*$ is derived using the forward process of a diffusion model, applied to $x_0^*$. Additionally, $\hat{x}_{t_n}^{*,\theta}$ is the result of one discretization step of an ODE solver applied to the PF-ODE %defined in Eq.~\eqref{eq_ODE_reverse}
, starting from $x_{t_{n+1}}^*$ and using a pre-trained diffusion model parameterized by $\theta$. Moreover, $d(\cdot, \cdot)$ is a distance measure, $\sigma$ is the sigmoid function, $n \sim \mathcal{U}(1, N )$, $N$ is the discretization length of the interval $[0, T ]$, and $\phi$
are the trainable parameters of the consistency model. We did not use a running average of $\phi$ for the Consistency-DPO loss because we already have the pre-trained model $f_{\mbox{\scriptsize{ref}}}$, which has the necessary self-consistency property.

\begin{algorithm*}[!t]
\caption{Curriculum DPO (for consistency models)}
\label{alg:method}
%\begin{algorithmic}
%\small{
\KwIn{$\{(x_{0, i}, c)\}_{i=1}^M$ - the training samples, $r_\varphi(x_0, c)$ - the reward model which can be conditioned on $c$, $B$ -  the number of batches for splitting the set of pairs, $\theta$ - the parameters of a pre-trained diffusion model, $\alpha_t, \sigma_t$ - the parameters of the noise schedule, $T$ - the last time step of diffusion, $N$ - the discretization length of the interval $[0, T]$, $\beta$ - DPO hyperparameter to control the divergence from the initial pre-trained state, $\sigma$ - the sigmoid function, $d^*$ - as defined in Eq.~\eqref{eq_d_star}, $\eta$ - the learning rate, $\{H_k\}_{k=1}^B$ - the number of training iterations after including the $k$-th batch.}
\KwOut{
$\phi$ - the trained weights of the generative model.
}
$\hat{X} \leftarrow \{(x_{0,i}, c)|r_\varphi(x_{0,i}, c) \leq r_\varphi(x_{0,i-1},  c), i=\{2,3,...,M\}\};$ \textcolor{commentGreen}{$\lhd$ sort the samples in descending order of the rewards}\\
$S \leftarrow \left\{(x_{0,i}, x_{0, j}, c)| i,j \in \{1, \dots M\}; i<j; x_{0,i}, x_{0, j} \in \hat{X},  r_\varphi(x_{0,i}, c) > r_\varphi(x_{0,j}, c)  \right\}$; \textcolor{commentGreen}{$\lhd$ create pairs of examples using the order from $\hat{X}$}\\
$\mathrm{L}_k \leftarrow \left\{\frac{(M-1) \cdot(B-k)}{B}\right\}_{k=1}^B$; \textcolor{commentGreen}{$\lhd$ the minimum preference limits of the batches}\\
$\mathrm{R}_k \leftarrow \left\{\frac{(M-1)\cdot(B-(k-1))}{B}\right\}_{k=1}^B$;  \textcolor{commentGreen}{$\lhd$ the maximum preference limits of the batches}\\
$S_k\leftarrow \left\{(x_{0}^w, x_{0}^l, c) | (x_{0}^w, x_{0}^l) = (x_{0,i}, x_{0, j}); \mathrm{L}_k   < j-i \leq \mathrm{R}_k  ; (x_{0,i}, x_{0, j}, c) \in S \right\}_{k = 1}^B$;  \textcolor{commentGreen}{$\lhd$ the batches of increasingly difficult pairs}\\
$P \leftarrow \emptyset$;  \textcolor{commentGreen}{$\lhd$ current training set} \\
\ForEach{$k \in \{1, \dots, B\}$}
{
    $P \leftarrow P \cup S_k$; \textcolor{commentGreen}{$\lhd$ include a new batch in the training}

    \ForEach{$i \in \{1, \dots, H_k\}$}
    {
    $(x_0^w, x_0^l, c) \sim \mathcal{U}(P)$; 
    $n \sim \mathcal{U}[1,N-1]$; $\epsilon \sim \mathcal{N}(0, \mathbf{I})$;\\
    $x_{t_{n+1}}^w \leftarrow \alpha_{t_{n+1}} x_0^w + \sigma_{t_{n+1}} \epsilon$;  \textcolor{commentGreen}{$\lhd$ forward process} \\
    $x_{t_{n+1}}^l \leftarrow \alpha_{t_{n+1}} x_0^l + \sigma_{t_{n+1}} \epsilon$; \textcolor{commentGreen}{$\lhd$ forward process}\\
    $\hat{x}_{t_{n}}^{w,\theta} \leftarrow \mathrm{ODESolver}(x_{t_{n+1}}^w, \theta, \alpha_{t_{n+1}}, \sigma_{t_{n+1}})$; \textcolor{commentGreen}{$\lhd$ one denoising step} \\
    $\hat{x}_{t_{n}}^{l,\theta} \leftarrow \mathrm{ODESolver}(x_{t_{n+1}}^l, \theta, \alpha_{t_{n+1}}, \sigma_{t_{n+1}})$; \textcolor{commentGreen}{$\lhd$ one denoising step}\\
    $\mathcal{L}_\mathrm{\mbox{\scriptsize{Con-DPO}}}(\phi) \leftarrow - \log \sigma\Big(-\beta( d^w(x_{t_{n+1}}^w,\hat{x}_{t_n}^{w,\theta}, \phi) - d^l(x_{t_{n+1}}^l,\hat{x}_{t_n}^{l,\theta}, \phi))\Big)$; \textcolor{commentGreen}{$\lhd$ DPO loss, as in Eq.~\eqref{eq_consistency_dpo}}\\
    $\phi \leftarrow \phi - \eta \frac{\partial \mathcal{L}_\mathrm{\mbox{\tiny{Con-DPO}}}}{\partial \phi}$; \textcolor{commentGreen}{$\lhd$ update the weights}
    }
    
}
\textbf{return} $\phi$
%}
%\end{algorithmic}
% \vspace*{-0.2cm}
\end{algorithm*}

To show that the Consistency-DPO loss defined in Eq.~\eqref{eq_consistency_dpo} has similar properties to the original DPO and Diffusion-DPO losses defined in Eq.~\eqref{eq_dpo} and Eq.~\eqref{eq_diffusion_dpo}, respectively, we can consider the gradient with respect to $\phi$:
\begin{equation}
    \label{eq_grad_consistency_dpo}
    \begin{split}
        \frac{\partial \mathcal{L}_\mathrm{\mbox{\scriptsize{Con-DPO}}}}{\partial \phi}\!=\!  \beta\mathbb{E}_{x_{t_{n+1}}^w\!, x_{t_{n+1}}^l\!, c}\!\Bigg[\!\sigma\!\left(\beta\!\left(d^w\!-\!d^l\right)\!\right)\!\!\left(\!\frac{\partial d^w}{ \partial \phi}\!-\!\frac{\partial d^l}{\partial \phi}\!\right)\!\!\!\Bigg]\!.
    \end{split}
\end{equation}
The gradient described in Eq.~\eqref{eq_grad_consistency_dpo} exhibits similar properties to the gradient outlined in Eq.~\eqref{eq_grad_dpo}. Specifically, it reduces the distance metric $d$ between $f_\phi$ and $f_{\mbox{\scriptsize{ref}}}$ for the favored examples and, at the same time, it increases the same metric for less preferred examples. Additionally, this gradient is also weighted by the sigmoid function, which is close to $1$ when $f_\phi$ shows a weaker preference for the favored example ($x_{t_{n+1}}^w$) than the level of preference yielded by $f_{\mbox{\scriptsize{ref}}}$.

\vspace{-0.1cm}
\subsection{Curriculum DPO}
\vspace{-0.1cm}
Despite the promising results of DPO, we conjecture that randomly sorting the available pairs $(x_0^w, x_0^l)$ without considering their difficulty during training is suboptimal. Our perspective is that the samples in each pair may be scored differently by a reward model due to a range of factors (style, realism, aesthetics, etc.), each with its own level of details and complexity. Consequently, we think that it would be advantageous for the generative model to initially encounter and learn the more apparent (coarse-level) factors that sit behind the preference judgment. As training progresses, we can gradually introduce pairs that differ by increasingly subtle (fine-level) details. Structuring the learning process through this strategy allows generative models to progressively refine their capabilities, leading to generated samples that are better aligned with the reward model. To this end, we propose a curriculum that utilizes the reward model $r_\varphi(x_0, c)$ to rank the available samples based on preference. Subsequently, we create pairs of samples, denoted as $(x_0^w, x_0^l)$, with varying levels of difficulty, leveraging the difference between the rank positions (preference scores) of the two samples as a measure of difficulty. Samples that are far apart in the preference ranking form easy pairs, under the assumption that the preferred sample is more easily identifiable thanks to obvious factors. Samples that are closely ranked present a greater challenge and are considered hard pairs, as distinguishing the preferred sample becomes less apparent. However, we set a minimum difference threshold in terms of preference, to make sure that the generative model does not learn preference patterns that are indistinguishable. This prevents the generative model from overfitting to the biases of the reward model or the noise in the data.

%\vspace{-0.1cm}
We formally describe Curriculum DPO for consistency models in Algorithm~\ref{alg:method}. In step 1, the set of examples $\{x_{0,i}\}_{i=1}^{M}$ that are generated under a condition $c$ (\ie~a 
text prompt) are sorted based on the preferences indicated by a reward model $r_\varphi$, such that the ranking follows $r_\varphi(x_{0,1}, c) \geq \dots \geq r_\varphi(x_{0,M}, c)$.
In a typical DPO setup, the pairs are formed from the set defined in step 2 of the algorithm, denoted by $S$. %=\left\{\!(x_{0,i}, x_{0,j}, c)| i,j\!\in\!\left\{1, \dots, M\right\}\!, i\!<\!j, r_\varphi(x_{0,i}, c)\!>\! r_\varphi(x_{0,j}^l, c)\!\right\}$. 
In steps 3-5, Curriculum DPO segments this set into $B$ distinct batches, where each batch is denoted by $S_k$, %= \left\{(x_{0}^w, x_{0}^l, c) | (x_{0}^w, x_{0}^l) = (x_{0,i}, x_{0, j}); \mathrm{L}_k   < j-i \leq \mathrm{R}_k  ; (x_{0,i}, x_{0, j}, c) \in S \right\}$, 
where $k = \{1,2,..., B\}$. Each batch $S_k$ (not to be confused with mini-batches) is defined by the limits $\mathrm{L}_k$ and $\mathrm{R}_k$, which represent the smallest and largest possible differences between positions $j$ and $i$ within the ranking made by the reward model. The training is executed in steps 7-16. At each iteration, the current batch $S_k$ is first included in the training set $P$. When $k=1$, the training set will comprise the first batch, $S_1$, which will be used to gradually adapt the generative model to the easiest (most distinctive) pairs, enhancing the training efficiency during the early learning iterations. After adding the current batch $S_k$ to the training set at step 8, the generative model is trained for $H_k$ iterations on $P$. The training iterations (steps 10-16) are custom to each specific DPO implementation, namely Diffusion-DPO and Consistency-DPO. In Algorithm~\ref{alg:method}, we present the iterations that are specific to the Consistency-DPO approach. In Appendix \ref{supp_alg}, we present the analogous version for Diffusion-DPO.

\vspace{-0.1cm}
\section{Experiments}
\label{sec: experiments}

\begin{table*}[t]

  \centering
  \setlength\tabcolsep{0.46em}
  \small{
  \begin{tabular}{cccccccc}
  \toprule
      &      & \multicolumn{3}{c}{Dataset $D_1$ \cite{Black-ICLR-2024}}   & \multicolumn{3}{c}{Dataset $D_2$ \cite{Saharia-NeurIPS-2022}} \\ 
    \cmidrule(l{2pt}r{2pt}){3-5}
    \cmidrule(l{2pt}r{2pt}){6-8}
    {Model}     & {Fine-Tuning Strategy}     & Text  & \multirow{2}{*}{Aesthetics} & Human   & Text & \multirow{2}{*}{Aesthetics} & Human  \\  
         &      &  Alignment &  & Preference  &  Alignment &  &  Preference \\
    \midrule
    \multirow{4}{*}{LCM} & - & 0.7243$_{\pm0.0048}$  & 6.0490$_{\pm0.0162}$ & 0.2912$_{\pm0.0021}$  & 0.5602$_{\pm0.0032}$ &  5.8038$_{\pm0.0139}$ & 0.2610$_{\pm0.0016}$ \\
    &  DDPO \cite{Black-ICLR-2024} & 0.7490$_{\pm0.0036}$ & 6.3730$_{\pm0.0130}$ & 0.2952$_{\pm0.0011}$ & 0.5721$_{\pm0.0043}$  & 6.0121$_{\pm0.0127}$  & 0.2803$_{\pm0.0019}$ \\

    &  DPO \cite{Wallace-arxiv-2023} & 0.7502$_{\pm 0.0045}$ & 6.4741$_{\pm0.0095}$ & 0.2990$_{\pm 0.0010}$ & 0.5720$_{\pm0.0040}$ & 6.0430$_{\pm0.0113}$ & 0.2814$_{\pm0.0023}$ \\

    & Curriculum DPO (ours) & \textbf{0.7548}$_{\pm0.0041}$ & \textbf{6.6417}$_{\pm0.0083}$& \textbf{0.3237}$_{\pm0.0012}$ & \textbf{0.5812}$_{\pm0.0038}$ & \textbf{6.1829}$_{\pm0.0128}$ & \textbf{0.2851}$_{\pm0.0017}$ \\

  \midrule
  \multirow{4}{*}{SD} & -  & 0.6804$_{\pm0.0052}$ & 5.5152$_{\pm0.0166}$ & 0.2784$_{\pm0.0015}$ & 0.5997$_{\pm0.0067}$ &  5.4292$_{\pm0.0181}$ &
  0.2646$_{\pm0.0011}$ \\
  & DDPO \cite{Black-ICLR-2024} & 0.7629$_{\pm0.0040}$ &
  5.8183$_{\pm0.0129}$ &
  0.2854$_{\pm 0.0012}$ &
  0.6024$_{\pm0.0055}$ &
  5.6748$_{\pm0.0162}$ &
  0.2673$_{\pm0.0025}$ \\
    
  &  DPO \cite{Wallace-arxiv-2023} & 0.7614$_{\pm0.0049}$ & 5.7146$_{\pm0.0162}$ & 0.2827$_{\pm0.0008}$ & 0.6075$_{\pm0.0057}$ &
  5.6205$_{\pm0.0152}$ &
  0.2672$_{\pm0.0013}$ \\

    & Curriculum DPO (ours) & \textbf{0.7703}$_{\pm0.0036}$ & \textbf{5.8232}$_{\pm0.0197}$ & \textbf{0.2856}$_{\pm0.0015}$ & \textbf{0.6234}$_{\pm0.0050}$ & \textbf{5.7060}$_{\pm0.0177}$ & \textbf{0.2681}$_{\pm0.0019}$ \\

  \bottomrule
  \end{tabular}
  }\vspace{-0.2cm}
    \caption{Text alignment, aesthetic and human preference scores on datasets $D_1$ and $D_2$, obtained by the baseline (pre-trained) LCM and SD models versus the three fine-tuning strategies: DDPO, DPO and Curriculum DPO. The results represent averages over three runs. The best scores are highlighted in bold.}
  \label{tab_strategies}
  \vspace{-0.3cm}
\end{table*}

\begin{figure*}[t]
  \centering
  \includegraphics[width=0.68\linewidth]{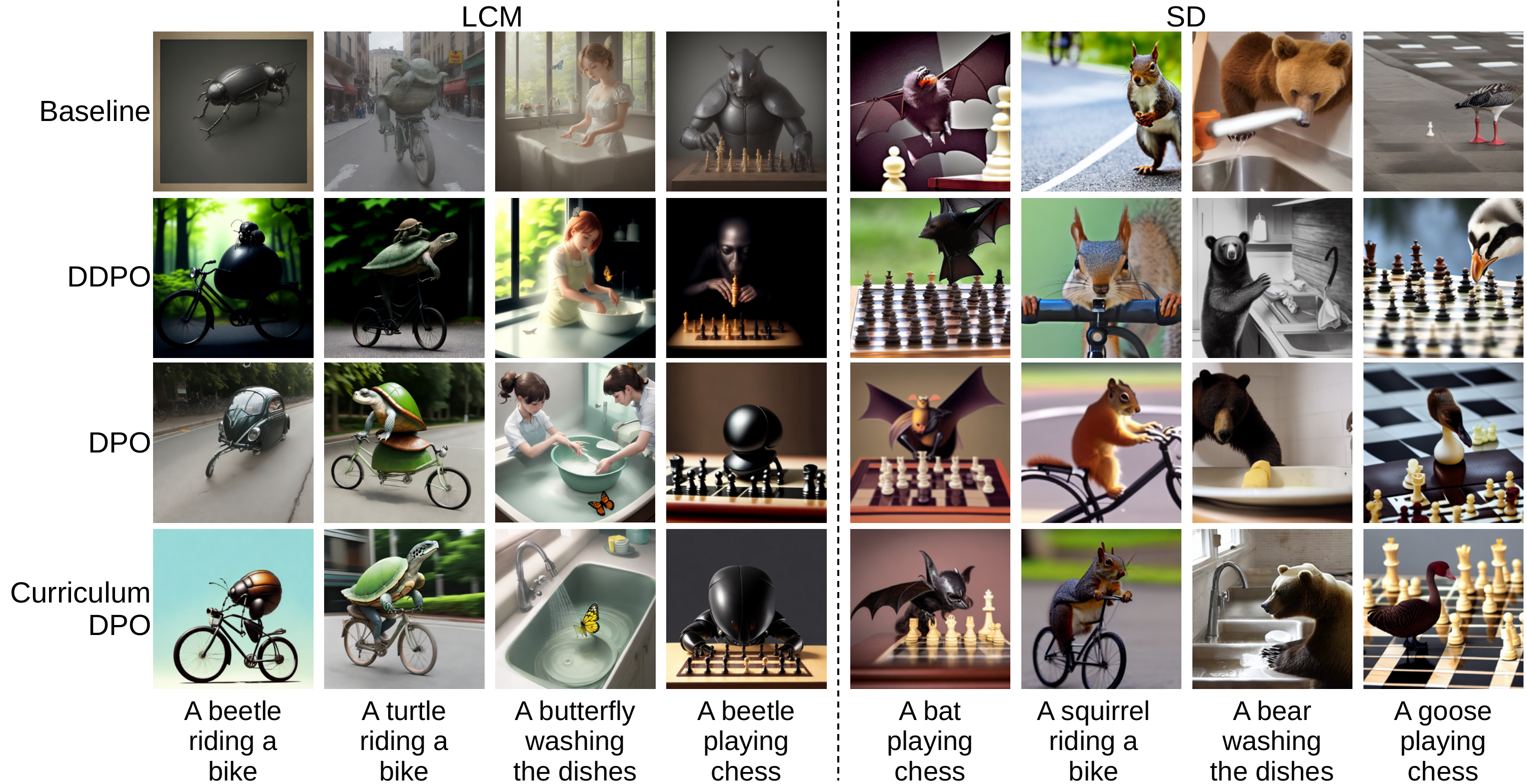}
  \vspace{-0.2cm}
   \caption{Qualitative results on dataset $D_1$, before and after fine-tuning for the text alignment task. The fine-tuning alternatives are: DDPO, DPO and Curriculum DPO. Best viewed in color.}
   \vspace{-0.4cm}
   \label{qualitative_text_align}
\end{figure*}

% \begin{table*}[t!]
%   \centering
%   \begin{tabular}{ccccc}
%   \toprule
%     Model     & Fine-Tuning Strategy     & Text Alignment & Aesthetics & Human Preference \\ 
%     \midrule
%     \multirow{4}{*}{LCM} & - \\
%     &  DDPO  \\

%     &  DPO \\

%     & Curriculum DPO (ours) \\
%     \midrule
%      \multirow{4}{*}{SD} & -  \\
%     &  DDPO  \\

%     &  DPO \\

%     & Curriculum DPO (ours)  \\

%   \bottomrule
%   \end{tabular}
  
%    \caption{Text alignment, aesthetic and human preference scores obtained on the DrawBench dataset by the baseline (pre-trained) LCM and SD models versus the three fine-tuning strategies: DDPO, DPO and Curriculum DPO. The best scores are highlighted in bold.}
%   \label{tab_strategies_drawbench}
%   \vspace{-0.3cm}
% \end{table*}

\vspace{-0.1cm}
\textbf{Datasets.} We run our experiments on three datasets. The first dataset ($D_1$) is created with the procedure of
% To create the training datasets for the text alignment, aesthetics and human preference tasks, we use the same strategy as 
\citet{Black-ICLR-2024}. For the text alignment and human preference tasks, the text prompts are based on the subject-verb-object (SVO) pattern, \eg~``a dog riding a bike''. Following \citet{Black-ICLR-2024}, we use a list of 45 animals and 3 types of activities. For the aesthetics task, the prompts are based on the ``a photo of \emph{<an object>}'' template, where the object represents one of the same 45 animal classes. We generate 500 images for each text prompt. For the text alignment and human preference tasks, there are 67,500 (prompt, image) pairs for training, and 6,750 pairs for evaluation. For the aesthetics task, there are 22,500 pairs for training and 2,250 for evaluation. 

The second dataset ($D_2$) is DrawBench~\cite{Saharia-NeurIPS-2022}, which consists of 200 diverse prompts. As for the first dataset, we generate 500 images per prompt. Therefore, the training set consists of 100,000 (prompt, image) pairs, while the test set comprises 10,000 pairs.

% \TODO{}
The third dataset ($D_3$) is a subset of 150,000 (prompt, image) pairs from Pick-a-Pic~\cite{Kirstain-NeurIPS-2023}. For $D_3$, the generated images are already provided. The test set contains 500 prompts. 

% \vspace{-0.1cm}
\noindent
\textbf{Generative models.} We conduct our experiments by using two pre-trained text-to-image generative models: Stable Diffusion (SD) v1.5~\cite{rombach-CVPR-2022} and Latent Consistency Model (LCM)~\cite{Luo-arXiv-2023}. The LCM checkpoint is trained using consistency distillation from an SD v1.5 checkpoint. To generate images with SD, we use the DDIM sampler with 50 steps. For LCM, we generate data using 8 steps of the Multistep Latent Consistency Sampling~\cite{Luo-arXiv-2023} procedure. Finally, we consider two different resolutions, $256\times256$ pixels for SD and $768\times768$ pixels for LCM, respectively.

% \vspace{-0.1cm}
\noindent
\textbf{Reward models}. To assess the alignment between a text prompt and a generated image, we use the cosine similarity to compare the embeddings produced by Sentence-BERT~\cite{reimers-EMNLP-2019} for the original prompt and the caption generated by LLaVA~\cite{liu-NeurIPS-2024}. To evaluate visual appeal (aesthetics), we use the LAION Aesthetics Predictor~\cite{Schuhmann-laion-2022}, a linear model applied on CLIP \cite{radford-ICML-2021} that is trained on human-rated images. %and predicts a score on a scale from $1$ to $10$. 
Finally, for human preference estimation, we employ HPSv2 \cite{Wu-arXiv-2023}, a CLIP model fine-tuned on a dataset of image pairs ranked by humans. The reward models are established according to \citet{Black-ICLR-2024}. In each scenario, DPO, DDPO and Curriculum DPO use the same reward models, ensuring a fair comparison. We present results with an additional reward model (Phi-3 \cite{abdin-arxiv-2024}) in Table \ref{tab_exp_phi3} from the supplementary.

% \vspace{-0.1cm}
\noindent
\textbf{Competing methods}.
We compare Curriculum DPO against two state-of-the-art fine-tuning methods, namely Direct Preference Optimization (DPO)~\cite{Rafailov-NeurIPS-2023, Wallace-arxiv-2023} and Denoising Diffusion Policy Optimization (DDPO)~\cite{Black-ICLR-2024}, as well as the original generative models, namely LCM and SD.

% \vspace{-0.1cm}
\noindent
\textbf{Training setup}. We conduct all experiments on a single A100 GPU, training each model for 10,000 training iterations, with a batch size of 16 and two steps of gradient accumulation. The training time is approximately 2 days for each LCM experiment, while using 64GB of VRAM. The SD experiments take approximately 1 day each, while using 36 GB of VRAM. We employ the AdamW optimizer with a constant learning rate of 
$3 \cdot 10^{-4}$. LoRA is activated in all experiments. For LCM, we utilize trainable low-rank matrices with a dimension of 64 and an equivalent $\alpha$. In the case of SD, we employ a rank of 8 and an $\alpha$ of 32. Following \citet{Wallace-arxiv-2023}, we set $\beta=5000$ for Diffusion-DPO. For Consistency-DPO, we set $\beta=200$ via grid search over the set $\{50, 100, 200, 300, 500 \}$. Similarly, we use grid search over the set $\{3, 5, 7\}$ to determine the number of curriculum batches for Curriculum DPO. The optimal value is $B=5$. The number of training iterations per batch is set to $K=400$ for the first four batches, \ie~$H_i=K, \forall i \in \{1,2,3,4\}$. For the last batch, the number of iterations is set to $H_5=10000-4\cdot K$. Hence, the total number of training iterations is the same as for the baseline, \ie~$\sum_{i=1}^B H_i=10000$.
Since all experiments involve reward models, the image pairs are sampled based on their preference score difference rather than their ranking difference, thus sidestepping the management of skewed preference score distributions.

% \vspace{-0.1cm}
\noindent
\textbf{Results.} For the quantitative assessment, we measure the reward scores after fine-tuning the models on each of the three tasks, corresponding to the three reward models. The results on datasets $D_1$ and $D_2$ are shown in Table~\ref{tab_strategies} (results on $D_3$ are presented in Table \ref{tab_exp_pickapic} from the supplementary). As anticipated, the original models yield the lowest scores, indicating that fine-tuning is generally useful. Both DPO and DDPO exhibit better performance than the baseline, but without a clear winner among the two approaches. In contrast, Curriculum DPO is the most effective method, achieving the best reward score on each task and each dataset. Moreover, the gains of Curriculum DPO are significant in most cases.

%\vspace{-0.1cm}
We present qualitative results for the text alignment task in Figure~\ref{qualitative_text_align}. Considering the examples from the first column, we observe that Curriculum DPO is the only method which generates the beetle and the bike in a coherent picture. Moreover, in the third column, our method is the only one able to generate the butterfly without additional humans around. Regarding the examples generated with SD, Curriculum DPO exhibits clear improvements over the other methods when generating the goose (last column) and the squirrel riding a bike (sixth column). Qualitative results for the aesthetics and human preference tasks are presented in Appendix \ref{supp_qual}.

% \begin{table*}[t]
%   \centering
%   \begin{tabular}{ccccc}
%   \toprule
%     \multirow{2}{*}{Fine-Tuning Strategy}     & \multicolumn{2}{c}{Text Alignment} & \multicolumn{2}{c}{Aesthetics} \\ 
%     & LCM & SD & LCM & SD \\
%     \midrule
%      - & 2.997 $\pm$ 0.274  & 2.438 $\pm$ 0.162
%      & 3.184 $\pm$ 0.438 & 3.136 $\pm$ 0.367\\
%     DDPO \cite{Black-ICLR-2024} & 2.878 $\pm$ 0.285 & 3.199 $\pm$ 0.165
%     & 2.819 $\pm$ 0.829 & 3.097 $\pm$ 0.463\\
%     DPO \cite{Wallace-arxiv-2023} & 3.041 $\pm$ 0.257 & 3.038 $\pm$ 0.268 
%  & 2.794 $\pm$ 0.301 & 3.056 $\pm$ 0.412\\

%     Curriculum DPO (ours) & \textbf{3.642} $\pm$ 0.373 & \textbf{3.394} $\pm$ 0.287
%  & \textbf{3.244} $\pm$ 0.587 & \textbf{3.384} $\pm$ 0.559\\
%   \bottomrule
%   \end{tabular}
%   \vspace{-0.3cm}
%    \caption{Average ratings of the baseline models and the three fine-tuning strategies (DDPO, DPO and Curriculum DPO) based on our human evaluation study. Each human annotated 1,280 randomly sampled images  with a rating from 1 to 5. The study was completed by 4 human evaluators. The best average ratings are highlighted in bold.}
%    \label{tab_human}
% \end{table*}

\begin{table}[t]
  \centering
    \setlength\tabcolsep{0.44em}
    \small{
  \begin{tabular}{ccccc}
  \toprule
    \multirow{2}{*}{Fine-Tuning Strategy}     & \multicolumn{2}{c}{Text Alignment} & \multicolumn{2}{c}{Aesthetics} \\ 
    & LCM & SD & LCM & SD \\
    \midrule
     - & 2.778 %$\pm$ 0.495
  & 2.276 %$\pm$ 0.321
     & 2.718 %$\pm$ 0.740
 & 3.510 \\%$\pm$ 0.654\\
    DDPO & 2.810 %$\pm$ 0.433
 & 2.983 %$\pm$ 0.456
 & 2.765 %$\pm$ 0.813
 & 3.454 %$\pm$ 0.647
\\
    DPO  & 2.846 %$\pm$ 0.543
 & 2.821 %$\pm$ 0.433
 & 2.782 %$\pm$ 0.804
 & 2.906 %$\pm$ 0.661
\\
    Curriculum DPO (ours) & \textbf{3.440} %$\pm$ 0.655
 & \textbf{3.175} %$\pm$ 0.458
 & \textbf{3.006} %$\pm$ 0.847
  & \textbf{3.664} %$\pm$ 0.674
\\
  \bottomrule
  \end{tabular}
  }
  \vspace{-0.2cm}
  \caption{Average ratings of the baseline models and the three fine-tuning strategies based on our human evaluation study. The study was completed by 9 human evaluators. The best average ratings are highlighted in bold. Statistical testing indicates that the voting results are statistically significant, with a p-value below 0.005.}
  \label{tab_human}
  \vspace{-0.4cm}
\end{table}

% \vspace{-0.1cm}

\begin{figure*}[!t]
\begin{subfigure}[t]{.245\textwidth}
  \centering
  % include second image
  \includegraphics[width=.99\linewidth]{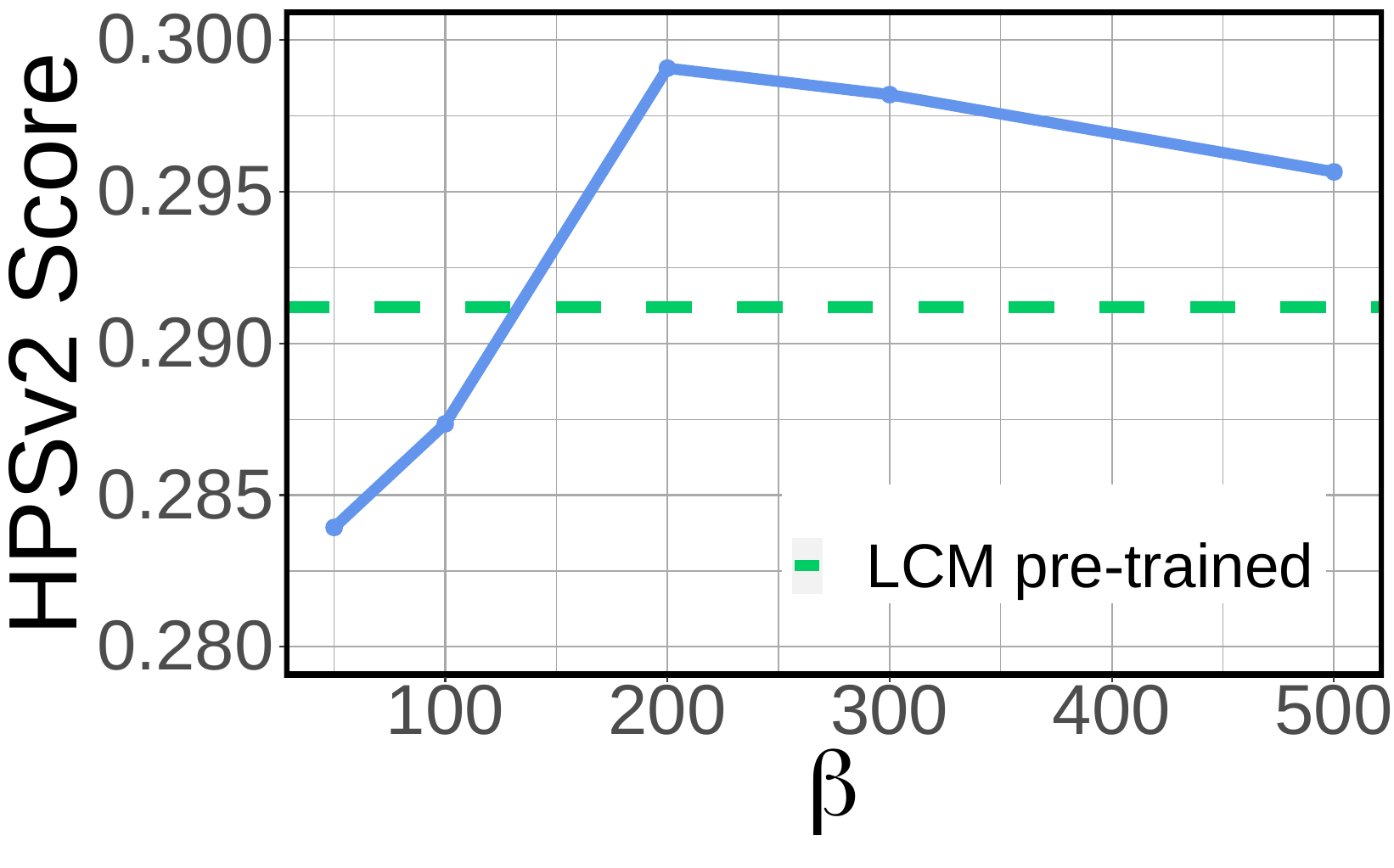}  
  \vspace{-0.5cm}
  \caption{Varying $\beta$ for Consistency-DPO.}
  \label{fig:sub-beta}
\end{subfigure}
\begin{subfigure}[t]{.245\textwidth}
  \centering
  % include third image
\includegraphics[width=.99\textwidth]{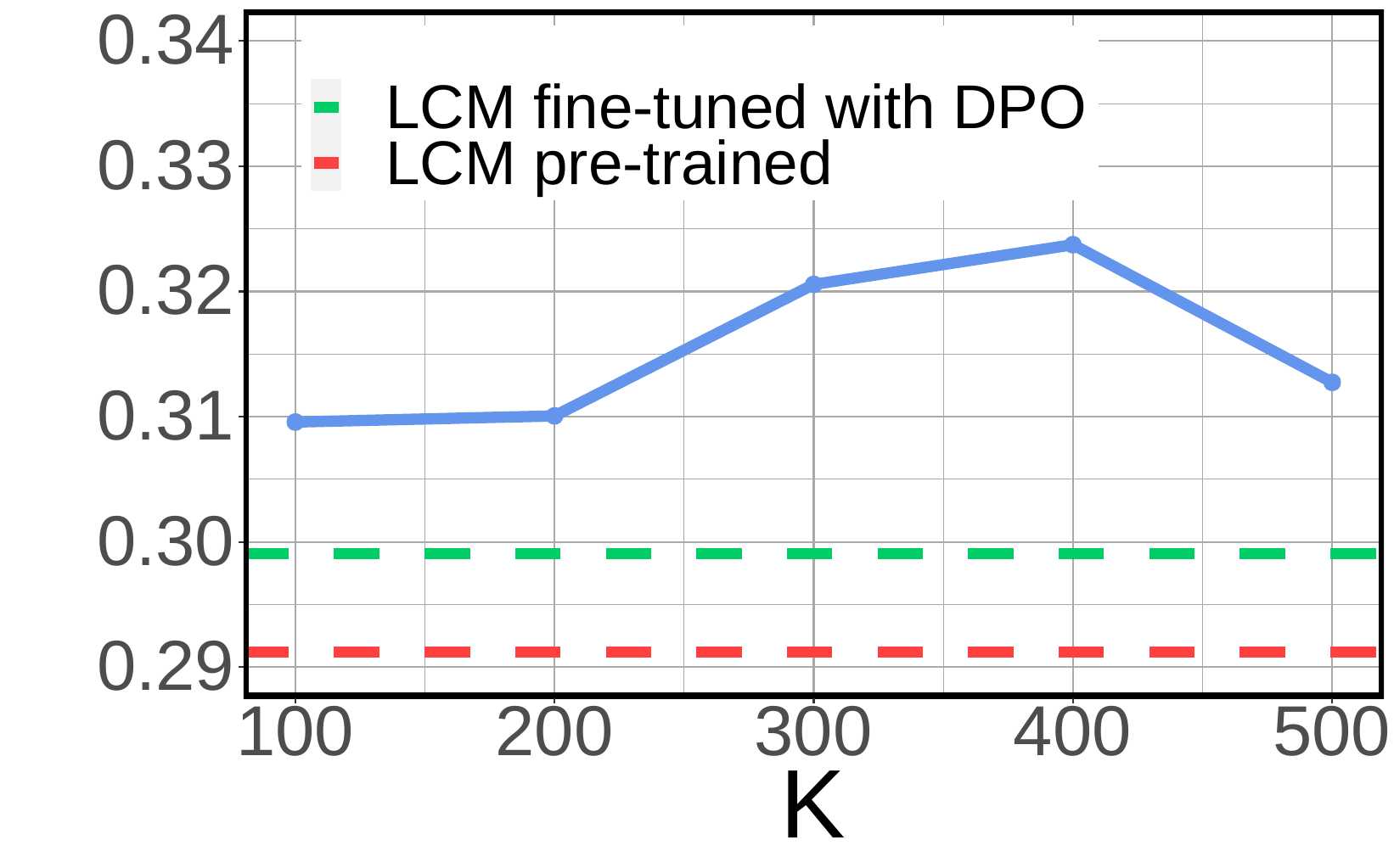}  
\vspace{-0.5cm}
  \caption{Varying $K$ for Curriculum DPO.}
  \label{fig:sub-K}
\end{subfigure}
\begin{subfigure}[t]{.245\textwidth}
  \centering
  % include fourth image
  \includegraphics[width=.99\linewidth]{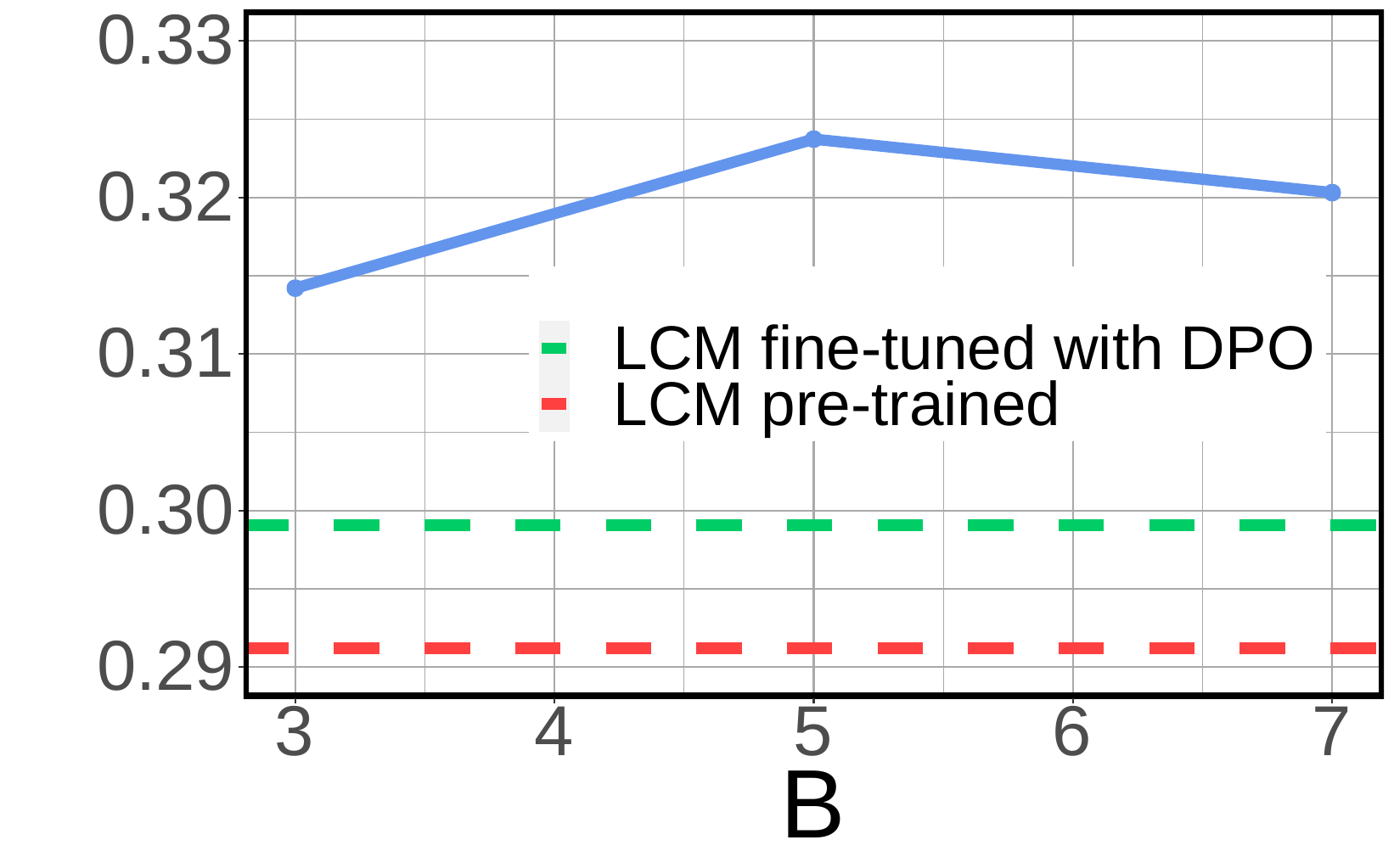} 
  \vspace{-0.5cm}
  \caption{Varying $B$ for Curriculum DPO.}
  \label{fig:sub-B}
\end{subfigure}
\begin{subfigure}[t]{.245\textwidth}
  \centering
  % include third image
\includegraphics[width=.99\textwidth]{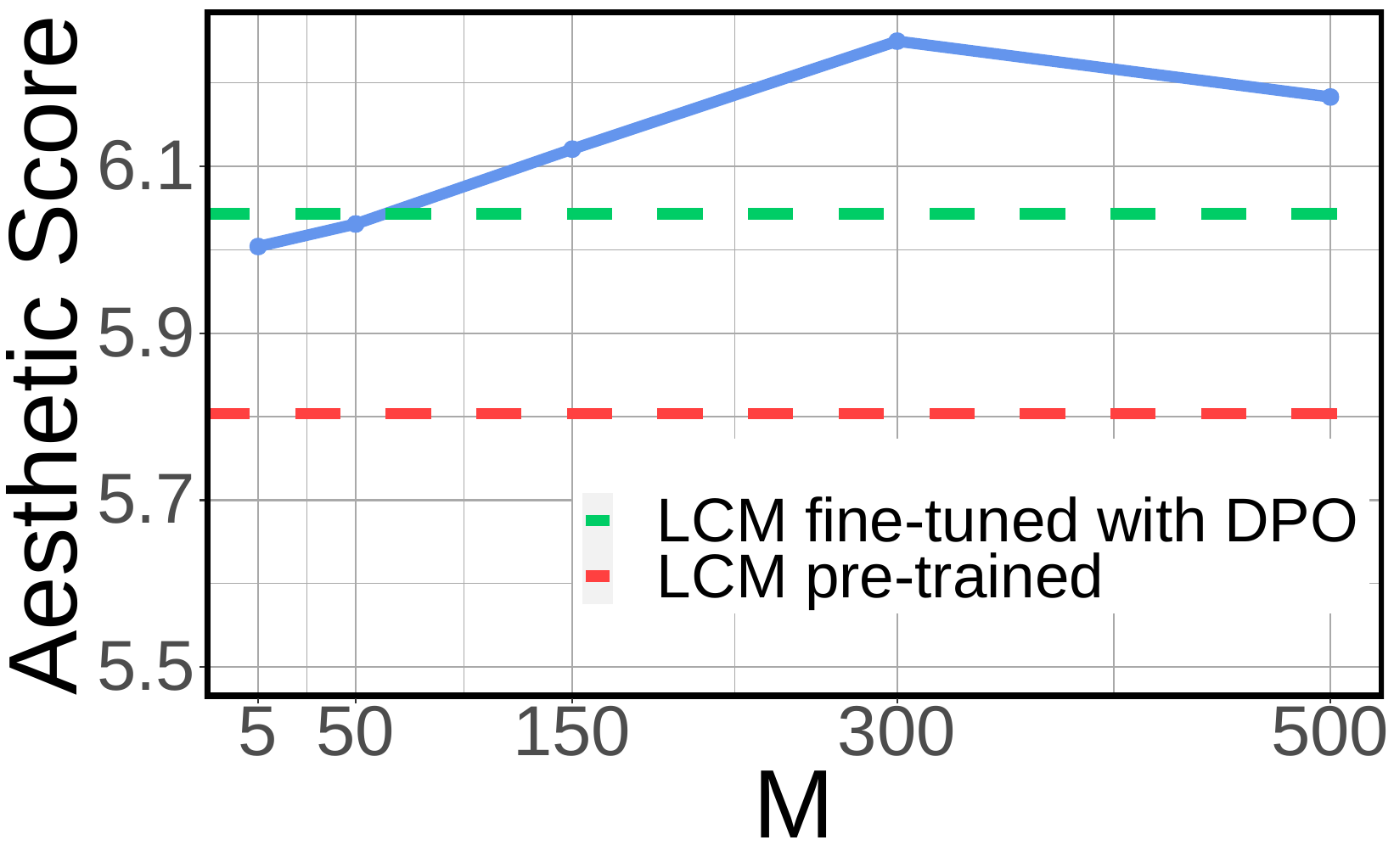}  
\vspace{-0.5cm}
  \caption{Varying $M$ for Curriculum DPO.}
  \label{fig:sub-M}
\end{subfigure}
\vspace{-0.25cm}
\caption{Ablation results obtained by varying the hyperparameter $\beta$ for Consistency-DPO (in {\color{RoyalBlue}blue}), the number of training iterations per batch $K$ for Curriculum DPO (in {\color{RoyalBlue}blue}), the number of batches $B$ for Curriculum DPO (in {\color{RoyalBlue}blue}), and the number of training images per prompt $M$ (in {\color{RoyalBlue}blue}). Fine-tuned LCM models are compared with the pre-trained LCM baseline on the human preference and visual appeal tasks (where scores are given by the HPSv2 reward model and LAION Aesthetics Predictor respectively).}
\vspace{-0.1cm}
\label{fig_ablation}
\end{figure*}

\begin{table*}[t]
  
  \centering
  \small{
  \begin{tabular}{cccccccc}
  \toprule
    {Model}     & {LoRA}     & {DPO} & Curriculum DPO & Text Alignment & {Aesthetics} & Human Preference \\ 
    %& & & DPO &  Alignment & & Preference \\
    \midrule
    \multirow{4}{*}{LCM} & \textcolor{Red}{\xmark} & \textcolor{Red}{\xmark} & \textcolor{Red}{\xmark} & 0.7243$_{\pm0.0048}$ & 6.0490$_{\pm0.0162}$ & 0.2912$_{\pm0.0021}$\\
    %\cmidrule(r){2-5}
    & \textcolor{ForestGreen}{\checkmark} & \textcolor{Red}{\xmark}& \textcolor{Red}{\xmark} & 0.7151$_{\pm0.0032}$ & 5.7158$_{\pm0.0154}$ & 0.2881$_{\pm0.0017}$\\
    %\cmidrule(r){2-5}
    &  \textcolor{ForestGreen}{\checkmark} &  \textcolor{ForestGreen}{\checkmark} & \textcolor{Red}{\xmark} & 0.7502$_{\pm0.0045}$ & 6.4741$_{\pm0.0095}$ & 0.2990$_{\pm0.0010}$\\
    %\cmidrule(r){2-5}
    &  \textcolor{ForestGreen}{\checkmark} & \textcolor{ForestGreen}{\checkmark} &  \textcolor{ForestGreen}{\checkmark} 
    & \textbf{0.7548}$_{\pm0.0041}$ & \textbf{6.6417}$_{\pm0.0083}$ & \textbf{0.3237}$_{\pm0.0012}$\\
    
  \midrule
  \multirow{4}{*}{SD} & \textcolor{Red}{\xmark} & \textcolor{Red}{\xmark} & \textcolor{Red}{\xmark} & 0.6804$_{\pm0.0052}$ & 5.5152$_{\pm0.0166}$ & 0.2784$_{\pm 0.0015}$\\
  %\cmidrule(r){2-5}
    
  &  \textcolor{ForestGreen}{\checkmark} & \textcolor{Red}{\xmark} &  \textcolor{Red}{\xmark} & 0.6852$_{\pm0.0047}$ & 5.6055$_{\pm0.0153}$ & 0.2720$_{\pm0.0013}$\\
  %    \cmidrule(r){2-5}
    
  &  \textcolor{ForestGreen}{\checkmark} &  \textcolor{ForestGreen}{\checkmark}& \textcolor{Red}{\xmark} & 0.7614$_{\pm0.0049}$ & 5.7146$_{\pm0.0162}$ & 0.2827$_{\pm0.0008}$\\

   % \cmidrule(r){2-5}
   & \textcolor{ForestGreen}{\checkmark} &  \textcolor{ForestGreen}{\checkmark} &   \textcolor{ForestGreen}{\checkmark}
   & \textbf{0.7703}$_{\pm0.0036}$ & \textbf{5.8232}$_{\pm0.0197}$ & \textbf{0.2856}$_{\pm0.0015}$ \\
  \bottomrule
  \end{tabular}
  }
  \vspace{-0.2cm}
  \caption{Ablation study on the key components of Curriculum DPO. LoRA, DPO and curriculum learning are gradually added to produce Curriculum DPO. The results represent averages over three runs. The best scores are highlighted in bold. Statistical significance testing for DPO versus Curriculum DPO attest that the differences brought by curriculum learning are significant (p-values are below 0.005).}
  \label{tab_ablation}
    \vspace{-0.3cm}
\end{table*}

\noindent
\textbf{Subjective human evaluation study.} We conducted a human evaluation study to evaluate the prompt alignment and aesthetics of the two pre-trained models, SD and LCM, along with their fine-tuned versions based on DDPO, DPO and Curriculum DPO, respectively. We asked the annotators to rate each image with a grade from $1$ to $5$, considering the task-specific evaluation criteria (prompt alignment / aesthetics). For each prompt, the generated images were shuffled and presented in a random order to prevent any form of cheating from the annotators. More details about the annotation process are provided in Appendix \ref{supp_human}. We present the average ratings in Table~\ref{tab_human}. We observe that Curriculum DPO outperforms the other methods by considerable margins, regardless of the task.

% \vspace{-0.1cm}
\noindent
\textbf{Ablation studies.} We conduct several ablation studies to explore the impact of various hyperparameters associated with Consistency-DPO and Curriculum DPO. In Figure~\ref{fig:sub-beta}, we show various values for $\beta$ and their impact on DPO when applied on consistency models. For values of $\beta$ higher or equal to 200, Consistency-DPO surpasses the pre-trained LCM baseline. In Figure~\ref{fig:sub-K}, we present an ablation on the impact of $K$ (the number of training iterations per curriculum batch) on the human preference scores returned by HPSv2. We observe better results for $K=300$ and $K=400$, although Curriculum DPO surpasses the baseline LCM for all values of $K$. In Figure~\ref{fig:sub-B}, we present the HPSv2 scores for various choices of the number of batches, $B$. Once again, Curriculum DPO consistently surpasses the pre-trained LCM, regardless of the number of training batches. In Figure~\ref{fig:sub-M}, we vary the number of generated images per prompt from $M=5$ to $M=500$. For reference, we include the results of DPO and DDPO based on $M=500$ images per prompt. Remarkably, Curriculum DPO is able to obtain similar scores to DPO and DDPO, while using 10$\times$ less training images. Moreover, Curriculum DPO outperforms the baseline LCM regardless of the number of training images.

%\vspace{-0.1cm}
Next, we present a detailed ablation study of the key components of Curriculum DPO in Table~\ref{tab_ablation}. This analysis showcases the impact of the LoRA fine-tuning technique, the DPO objective, and the curriculum strategy. Interestingly, applying LoRA to LCM degrades performance on all three tasks. However, combining LoRA with DPO or Curriculum DPO leads to significantly better results. Curriculum DPO consistently outperforms its ablated version, justifying the proposed design.

\vspace{-0.1cm}
\section{Conclusion}
\label{sec: conclusion}
\vspace{-0.1cm}

In this paper, we proposed a novel training strategy for diffusion and consistency models, which is based on introducing curriculum learning into Direct Preference Optimization. Furthermore, we integrated DPO into a recent category of efficient diffusion models, known as consistency models. Through comprehensive experiments consisting of three distinct tasks, we demonstrated significant performance gains due to our progressive training based on easy-to-hard image pairs. Aside from presenting numerical results on representative automatic metrics to assess our method and determine its benefits, we also carried out a subjective human evaluation study. The human evaluation study confirmed our observations, namely that curriculum learning brings significant performance gains when integrated into DPO. We thus conclude that our contribution leads to the development of stronger generative models that can accurately capture and synthesize human-preferred images. % In future work, we aim to extend and apply Curriculum DPO to LLMs.

\noindent
\textbf{Acknowledgments.}
This work was supported by a grant of the Ministry of Research, Innovation and Digitization, CCCDI - UEFISCDI, project number PN-IV-P6-6.3-SOL-2024-2-0227, within PNCDI IV.

{
    \small
    \bibliographystyle{ieeenat_fullname}
    \bibliography{references}

\begin{thebibliography}{79}
\providecommand{\natexlab}[1]{#1}
\providecommand{\url}[1]{\texttt{#1}}
\expandafter\ifx\csname urlstyle\endcsname\relax
  \providecommand{\doi}[1]{doi: #1}\else
  \providecommand{\doi}{doi: \begingroup \urlstyle{rm}\Url}\fi

\bibitem[Abdin et~al.(2024)Abdin, Aneja, Awadalla, Awadallah, Awan, Bach,
  Bahree, Bakhtiari, Bao, Behl, et~al.]{abdin-arxiv-2024}
Marah Abdin, Jyoti Aneja, Hany Awadalla, Ahmed Awadallah, Ammar~Ahmad Awan,
  Nguyen Bach, Amit Bahree, Arash Bakhtiari, Jianmin Bao, Harkirat Behl, et~al.
\newblock {Phi-3 technical report: A highly capable language model locally on
  your phone}.
\newblock \emph{arXiv preprint arXiv:2404.14219}, 2024.

\bibitem[Anderson(1982)]{Andreson-SPA-1982}
Brian~D.O. Anderson.
\newblock Reverse-time diffusion equation models.
\newblock \emph{Stochastic Processes and their Applications}, 12\penalty0
  (3):\penalty0 313--326, 1982.

\bibitem[Avrahami et~al.(2022)Avrahami, Lischinski, and
  Fried]{avrahami-CVPR-2022}
Omri Avrahami, Dani Lischinski, and Ohad Fried.
\newblock Blended diffusion for text-driven editing of natural images.
\newblock In \emph{Proceedings of CVPR}, pages 18208--18218, 2022.

\bibitem[Bai et~al.(2022)Bai, Kadavath, Kundu, Askell, Kernion, Jones, Chen,
  Goldie, Mirhoseini, McKinnon, et~al.]{bai-arXiv-2022}
Yuntao Bai, Saurav Kadavath, Sandipan Kundu, Amanda Askell, Jackson Kernion,
  Andy Jones, Anna Chen, Anna Goldie, Azalia Mirhoseini, Cameron McKinnon,
  et~al.
\newblock {Constitutional AI: Harmlessness from AI feedback}.
\newblock \emph{arXiv preprint arXiv:2212.08073}, 2022.

\bibitem[Baranchuk et~al.(2022)Baranchuk, Rubachev, Voynov, Khrulkov, and
  Babenko]{baranchuk-arXiv-2021}
Dmitry Baranchuk, Ivan Rubachev, Andrey Voynov, Valentin Khrulkov, and Artem
  Babenko.
\newblock {Label-Efficient Semantic Segmentation with Diffusion Models}.
\newblock In \emph{Proceedings of ICLR}, 2022.

\bibitem[Bengio et~al.(2009)Bengio, Louradour, Collobert, and
  Weston]{Bengio-ICML-2009}
Yoshua Bengio, J\'{e}r\^{o}me Louradour, Ronan Collobert, and Jason Weston.
\newblock {Curriculum Learning}.
\newblock In \emph{Proceedings of ICML}, pages 41--48, 2009.

\bibitem[Black et~al.(2024)Black, Janner, Du, Kostrikov, and
  Levine]{Black-ICLR-2024}
Kevin Black, Michael Janner, Yilun Du, Ilya Kostrikov, and Sergey Levine.
\newblock {Training Diffusion Models with Reinforcement Learning}.
\newblock In \emph{Proceedings of ICLR}, 2024.

\bibitem[Chao et~al.(2022)Chao, Sun, Cheng, Lo, Chang, Liu, Chang, Chen, and
  Lee]{chao-ICLR-2022}
Chen-Hao Chao, Wei-Fang Sun, Bo-Wun Cheng, Yi-Chen Lo, Chia-Che Chang, Yu-Lun
  Liu, Yu-Lin Chang, Chia-Ping Chen, and Chun-Yi Lee.
\newblock {Denoising Likelihood Score Matching for Conditional Score-Based Data
  Generation}.
\newblock In \emph{Proceedings of ICLR}, 2022.

\bibitem[Christiano et~al.(2017)Christiano, Leike, Brown, Martic, Legg, and
  Amodei]{Christiano-NeurIPS-2017}
Paul~F. Christiano, Jan Leike, Tom Brown, Miljan Martic, Shane Legg, and Dario
  Amodei.
\newblock {Deep Reinforcement Learning from Human Preferences}.
\newblock In \emph{Proceedings of NeurIPS}, pages 4302--4310, 2017.

\bibitem[Chung and Ye(2022)]{chung-MIA-2022}
Hyungjin Chung and Jong~Chul Ye.
\newblock {Score-based diffusion models for accelerated MRI}.
\newblock \emph{Medical Image Analysis}, 80:\penalty0 102479, 2022.

\bibitem[Croitoru et~al.(2023)Croitoru, Hondru, Ionescu, and
  Shah]{Croitoru-TPAMI-2023}
Florinel-Alin Croitoru, Vlad Hondru, Radu~Tudor Ionescu, and Mubarak Shah.
\newblock Diffusion models in vision: A survey.
\newblock \emph{IEEE Transactions on Pattern Analysis and Machine
  Intelligence}, 45:\penalty0 10850--10869, 2023.

\bibitem[Daniels et~al.(2021)Daniels, Maunu, and Hand]{daniels-NeurIPS-2021}
Max Daniels, Tyler Maunu, and Paul Hand.
\newblock Score-based generative neural networks for large-scale optimal
  transport.
\newblock In \emph{Proceedings of NeurIPS}, pages 12955--12965, 2021.

\bibitem[Dhariwal and Nichol(2021)]{dhariwal-NeurIPS-2021}
Prafulla Dhariwal and Alexander Nichol.
\newblock {Diffusion models beat GANs on image synthesis}.
\newblock In \emph{Proceedings of NeurIPS}, pages 8780--8794, 2021.

\bibitem[Doan et~al.(2019)Doan, Monteiro, Albuquerque, Mazoure, Durand, Pineau,
  and Hjelm]{doan-AAAI-2019}
Thang Doan, Joao Monteiro, Isabela Albuquerque, Bogdan Mazoure, Audrey Durand,
  Joelle Pineau, and R.~Devon Hjelm.
\newblock {On-line adaptative curriculum learning for GANs}.
\newblock In \emph{Proceedings of AAAI}, pages 3470--3477, 2019.

\bibitem[Fan et~al.(2023)Fan, Watkins, Du, Liu, Ryu, Boutilier, Abbeel,
  Ghavamzadeh, Lee, and Lee]{Fan-NeurIPS-2023}
Ying Fan, Olivia Watkins, Yuqing Du, Hao Liu, Moonkyung Ryu, Craig Boutilier,
  Pieter Abbeel, Mohammad Ghavamzadeh, Kangwook Lee, and Kimin Lee.
\newblock {DPOK: Reinforcement Learning for Fine-tuning Text-to-Image Diffusion
  Models}.
\newblock In \emph{Proceedings of NeurIPS}, pages 79858--79885, 2023.

\bibitem[Ghasedi et~al.(2019)Ghasedi, Wang, Deng, and Huang]{ghasedi-CVPR-2019}
Kamran Ghasedi, Xiaoqian Wang, Cheng Deng, and Heng Huang.
\newblock Balanced self-paced learning for generative adversarial clustering
  network.
\newblock In \emph{Proceedings of CVPR}, pages 4391--4400, 2019.

\bibitem[Gu et~al.(2022)Gu, Chen, Bao, Wen, Zhang, Chen, Yuan, and
  Guo]{gu-CVPR-2022}
Shuyang Gu, Dong Chen, Jianmin Bao, Fang Wen, Bo Zhang, Dongdong Chen, Lu Yuan,
  and Baining Guo.
\newblock Vector quantized diffusion model for text-to-image synthesis.
\newblock In \emph{Proceedings of CVPR}, pages 10696--10706, 2022.

\bibitem[Havrilla et~al.(2024)Havrilla, Du, Raparthy, Nalmpantis, Dwivedi-Yu,
  Zhuravinskyi, Hambro, Sukhbaatar, and Raileanu]{havrilla-arXiv-2024}
Alex Havrilla, Yuqing Du, Sharath~Chandra Raparthy, Christoforos Nalmpantis,
  Jane Dwivedi-Yu, Maksym Zhuravinskyi, Eric Hambro, Sainbayar Sukhbaatar, and
  Roberta Raileanu.
\newblock Teaching large language models to reason with reinforcement learning.
\newblock \emph{arXiv preprint arXiv:2403.04642}, 2024.

\bibitem[Ho and Salimans(2021)]{ho-NeurIPS-2021}
Jonathan Ho and Tim Salimans.
\newblock {Classifier-Free Diffusion Guidance}.
\newblock In \emph{Proceedings of NeurIPS Workshop on DGMs and Applications},
  2021.

\bibitem[Ho et~al.(2020)Ho, Jain, and Abbeel]{ho-NeurIPS-2020}
Jonathan Ho, Ajay Jain, and Pieter Abbeel.
\newblock Denoising diffusion probabilistic models.
\newblock In \emph{Proceedings of NeurIPS}, pages 6840--6851, 2020.

\bibitem[Ho et~al.(2022)Ho, Saharia, Chan, Fleet, Norouzi, and
  Salimans]{ho-arXiv-2021}
Jonathan Ho, Chitwan Saharia, William Chan, David~J. Fleet, Mohammad Norouzi,
  and Tim Salimans.
\newblock {Cascaded Diffusion Models for High Fidelity Image Generation}.
\newblock \emph{Journal of Machine Learning Research}, 23\penalty0
  (47):\penalty0 1--33, 2022.

\bibitem[Hu et~al.(2022)Hu, Wallis, Allen-Zhu, Li, Wang, Wang, Chen,
  et~al.]{Hu-ICLR-2022}
Edward~J. Hu, Phillip Wallis, Zeyuan Allen-Zhu, Yuanzhi Li, Shean Wang, Lu
  Wang, Weizhu Chen, et~al.
\newblock {LoRA: Low-Rank Adaptation of Large Language Models}.
\newblock In \emph{Proceedings of ICLR}, 2022.

\bibitem[Jiang et~al.(2015)Jiang, Meng, Zhao, Shan, and
  Hauptmann]{jiang-AAAI-2015}
Lu Jiang, Deyu Meng, Qian Zhao, Shiguang Shan, and Alexander Hauptmann.
\newblock Self-paced curriculum learning.
\newblock In \emph{Proceedings of AAAI}, pages 2694--2700, 2015.

\bibitem[Jim{\'e}nez-S{\'a}nchez et~al.(2019)Jim{\'e}nez-S{\'a}nchez, Mateus,
  Kirchhoff, Kirchhoff, Biberthaler, Navab, Gonz{\'a}lez~Ballester, and
  Piella]{Jimenez-MICCAI-2019}
Amelia Jim{\'e}nez-S{\'a}nchez, Diana Mateus, Sonja Kirchhoff, Chlodwig
  Kirchhoff, Peter Biberthaler, Nassir Navab, Miguel~A. Gonz{\'a}lez~Ballester,
  and Gemma Piella.
\newblock {Medical-based Deep Curriculum Learning for Improved Fracture
  Classification}.
\newblock In \emph{Proceedings of MICCAI}, pages 694--702, 2019.

\bibitem[Karras et~al.(2018)Karras, Aila, Laine, and
  Lehtinen]{karras-ICLR-2018}
Tero Karras, Timo Aila, Samuli Laine, and Jaakko Lehtinen.
\newblock Progressive growing of {GAN}s for improved quality, stability, and
  variation.
\newblock In \emph{Proceedings of ICLR}, 2018.

\bibitem[Karras et~al.(2020)Karras, Laine, Aittala, Hellsten, Lehtinen, and
  Aila]{karras-CVPR-2020}
Tero Karras, Samuli Laine, Miika Aittala, Janne Hellsten, Jaakko Lehtinen, and
  Timo Aila.
\newblock {Analyzing and improving the image quality of StyleGAN}.
\newblock In \emph{Proceedings of CVPR}, pages 8110--8119, 2020.

\bibitem[Kim et~al.(2025)Kim, Go, Kwon, and Kim]{kim-arXiv-2024}
Jin-Young Kim, Hyojun Go, Soonwoo Kwon, and Hyun-Gyoon Kim.
\newblock Denoising task difficulty-based curriculum for training diffusion
  models.
\newblock In \emph{Proceedings of ICLR}, 2025.

\bibitem[Kingma et~al.(2021)Kingma, Salimans, Poole, and
  Ho]{kingma-NeurIPS-2021}
Diederik Kingma, Tim Salimans, Ben Poole, and Jonathan Ho.
\newblock Variational diffusion models.
\newblock In \emph{Proceedings of NeurIPS}, pages 21696--21707, 2021.

\bibitem[Kirstain et~al.(2023)Kirstain, Polyak, Singer, Matiana, Penna, and
  Levy]{Kirstain-NeurIPS-2023}
Yuval Kirstain, Adam Polyak, Uriel Singer, Shahbuland Matiana, Joe Penna, and
  Omer Levy.
\newblock Pick-a-pic: An open dataset of user preferences for text-to-image
  generation.
\newblock In \emph{Proceedings of NeurIPS}, 2023.

\bibitem[Kumar et~al.(2010)Kumar, Packer, and Koller]{kuman-NeurIPS-2010}
M. Kumar, Benjamin Packer, and Daphne Koller.
\newblock Self-paced learning for latent variable models.
\newblock In \emph{Proceedings of NeurIPS}, pages 1189--1197, 2010.

\bibitem[Lee et~al.(2023)Lee, Liu, Ryu, Watkins, Du, Boutilier, Abbeel,
  Ghavamzadeh, and Gu]{lee-ArXiv-2023}
Kimin Lee, Hao Liu, Moonkyung Ryu, Olivia Watkins, Yuqing Du, Craig Boutilier,
  Pieter Abbeel, Mohammad Ghavamzadeh, and Shixiang~Shane Gu.
\newblock Aligning text-to-image models using human feedback.
\newblock \emph{arXiv preprint arXiv:2302.12192}, 2023.

\bibitem[Liu et~al.(2018)Liu, He, Liu, and Zhao]{Liu-IJCAI-2018}
Cao Liu, Shizhu He, Kang Liu, and Jun Zhao.
\newblock {Curriculum Learning for Natural Answer Generation}.
\newblock In \emph{Proceedings of IJCAI}, pages 4223--4229, 2018.

\bibitem[Liu et~al.(2023)Liu, Li, Wu, and Lee]{liu-NeurIPS-2024}
Haotian Liu, Chunyuan Li, Qingyang Wu, and Yong~Jae Lee.
\newblock Visual instruction tuning.
\newblock In \emph{Proceedings of NeurIPS}, pages 34892--34916, 2023.

\bibitem[Liu et~al.(2022)Liu, Ren, Lin, and Zhao]{liu-ICLR-2022}
Luping Liu, Yi Ren, Zhijie Lin, and Zhou Zhao.
\newblock {Pseudo Numerical Methods for Diffusion Models on Manifolds}.
\newblock In \emph{Proceedings of ICLR}, 2022.

\bibitem[Lugmayr et~al.(2022)Lugmayr, Danelljan, Romero, Yu, Timofte, and
  Van~Gool]{lugmayr-CVPR-2022}
Andreas Lugmayr, Martin Danelljan, Andres Romero, Fisher Yu, Radu Timofte, and
  Luc Van~Gool.
\newblock {RePaint: Inpainting using Denoising Diffusion Probabilistic Models}.
\newblock In \emph{Proceedings of CVPR}, pages 11461--11471, 2022.

\bibitem[Luo et~al.(2023{\natexlab{a}})Luo, Tan, Huang, Li, and
  Zhao]{Luo-arXiv-2023}
Simian Luo, Yiqin Tan, Longbo Huang, Jian Li, and Hang Zhao.
\newblock Latent consistency models: Synthesizing high-resolution images with
  few-step inference.
\newblock \emph{arXiv preprint arXiv:2310.04378}, 2023{\natexlab{a}}.

\bibitem[Luo et~al.(2023{\natexlab{b}})Luo, Tan, Patil, Gu, von Platen, Passos,
  Huang, Li, and Zhao]{Luo-Arxiv-2023b}
Simian Luo, Yiqin Tan, Suraj Patil, Daniel Gu, Patrick von Platen,
  Apolin{\'a}rio Passos, Longbo Huang, Jian Li, and Hang Zhao.
\newblock {LCM-LoRA: A Universal Stable-Diffusion Acceleration Module}.
\newblock \emph{arXiv preprint arXiv:2311.05556}, 2023{\natexlab{b}}.

\bibitem[Madan et~al.(2024)Madan, Ristea, Nasrollahi, Moeslund, and
  Ionescu]{Madan-WACV-2024}
Neelu Madan, Nicolae-C{\u{a}}t{\u{a}}lin Ristea, Kamal Nasrollahi, Thomas~B.
  Moeslund, and Radu~Tudor Ionescu.
\newblock {CL-MAE: Curriculum-Learned Masked Autoencoders}.
\newblock In \emph{Proceedings of WACV}, pages 2492--2502, 2024.

\bibitem[Meng et~al.(2021)Meng, Song, Song, Wu, Zhu, and
  Ermon]{meng-arXiv-2021}
Chenlin Meng, Yang Song, Jiaming Song, Jiajun Wu, Jun-Yan Zhu, and Stefano
  Ermon.
\newblock {SDEdit: Guided Image Synthesis and Editing with Stochastic
  Differential Equations}.
\newblock In \emph{Proceedings of ICLR}, 2021.

\bibitem[Morerio et~al.(2017)Morerio, Cavazza, Volpi, Vidal, and
  Murino]{morerio-ICCV-2017}
Pietro Morerio, Jacopo Cavazza, Riccardo Volpi, Ren{\'e} Vidal, and Vittorio
  Murino.
\newblock Curriculum dropout.
\newblock In \emph{Proceedings of ICCV}, pages 3544--3552, 2017.

\bibitem[Nichol and Dhariwal(2021)]{nichol-ICML-2021}
Alexander~Quinn Nichol and Prafulla Dhariwal.
\newblock Improved denoising diffusion probabilistic models.
\newblock In \emph{Proceedings of ICML}, pages 8162--8171, 2021.

\bibitem[Nichol et~al.(2022)Nichol, Dhariwal, Ramesh, Shyam, Mishkin, Mcgrew,
  Sutskever, and Chen]{nichol-ICML-2022}
Alexander~Quinn Nichol, Prafulla Dhariwal, Aditya Ramesh, Pranav Shyam, Pamela
  Mishkin, Bob Mcgrew, Ilya Sutskever, and Mark Chen.
\newblock {GLIDE: Towards Photorealistic Image Generation and Editing with
  Text-Guided Diffusion Models}.
\newblock In \emph{Proceedings of ICML}, pages 16784--16804, 2022.

\bibitem[Ouyang et~al.(2022)Ouyang, Wu, Jiang, Almeida, Wainwright, Mishkin,
  Zhang, Agarwal, Slama, Ray, et~al.]{ouyang-NeurIPS-2022}
Long Ouyang, Jeffrey Wu, Xu Jiang, Diogo Almeida, Carroll Wainwright, Pamela
  Mishkin, Chong Zhang, Sandhini Agarwal, Katarina Slama, Alex Ray, et~al.
\newblock Training language models to follow instructions with human feedback.
\newblock In \emph{Proceedings of NeurIPS}, pages 27730--27744, 2022.

\bibitem[Papineni et~al.(2002)Papineni, Roukos, Ward, and
  Zhu]{papineni-ACL-2002}
Kishore Papineni, Salim Roukos, Todd Ward, and Wei-Jing Zhu.
\newblock {BLEU: A Method for Automatic Evaluation of Machine Translation}.
\newblock In \emph{Proceedings of ACL}, pages 311--318, 2002.

\bibitem[Peng et~al.(2019)Peng, Kumar, Zhang, and Levine]{Peng-arXiv-2019}
Xue~Bin Peng, Aviral Kumar, Grace Zhang, and Sergey Levine.
\newblock {Advantage-Weighted Regression: Simple and Scalable Off-Policy
  Reinforcement Learning}.
\newblock \emph{arXiv preprint arXiv:1910.00177}, 2019.

\bibitem[Peters and Schaal(2007)]{Peters-ICML-2007}
Jan Peters and Stefan Schaal.
\newblock Reinforcement learning by reward-weighted regression for operational
  space control.
\newblock In \emph{Proceedings of ICML}, page 745–750, 2007.

\bibitem[Radford et~al.(2021)Radford, Kim, Hallacy, Ramesh, Goh, Agarwal,
  Sastry, Askell, Mishkin, Clark, Krueger, and Sutskever]{radford-ICML-2021}
Alec Radford, Jong~Wook Kim, Chris Hallacy, Aditya Ramesh, Gabriel Goh,
  Sandhini Agarwal, Girish Sastry, Amanda Askell, Pamela Mishkin, Jack Clark,
  Gretchen Krueger, and Ilya Sutskever.
\newblock Learning transferable visual models from natural language
  supervision.
\newblock In \emph{Proceedings of ICML}, pages 8748--8763, 2021.

\bibitem[Rafailov et~al.(2023)Rafailov, Sharma, Mitchell, Manning, Ermon, and
  Finn]{Rafailov-NeurIPS-2023}
Rafael Rafailov, Archit Sharma, Eric Mitchell, Christopher~D. Manning, Stefano
  Ermon, and Chelsea Finn.
\newblock {Direct Preference Optimization: Your Language Model is Secretly a
  Reward Model}.
\newblock In \emph{Proceedings of NeurIPS}, pages 53728--53741, 2023.

\bibitem[Ramesh et~al.(2022)Ramesh, Dhariwal, Nichol, Chu, and
  Chen]{ramesh-arXiv-2022}
Aditya Ramesh, Prafulla Dhariwal, Alex Nichol, Casey Chu, and Mark Chen.
\newblock {Hierarchical text-conditional image generation with CLIP latents}.
\newblock \emph{arXiv preprint arXiv:2204.06125}, 2022.

\bibitem[Reimers and Gurevych(2019)]{reimers-EMNLP-2019}
Nils Reimers and Iryna Gurevych.
\newblock {Sentence-{BERT}: Sentence Embeddings using {S}iamese
  {BERT}-Networks}.
\newblock In \emph{Proceedings of EMNLP}, pages 3982--3992, 2019.

\bibitem[Rombach et~al.(2022{\natexlab{a}})Rombach, Blattmann, Lorenz, Esser,
  and Ommer]{rombach-CVPR-2022}
Robin Rombach, Andreas Blattmann, Dominik Lorenz, Patrick Esser, and Bj{\"o}rn
  Ommer.
\newblock {High-Resolution Image Synthesis with Latent Diffusion Models}.
\newblock In \emph{Proceedings of CVPR}, pages 10684--10695,
  2022{\natexlab{a}}.

\bibitem[Rombach et~al.(2022{\natexlab{b}})Rombach, Blattmann, and
  Ommer]{rombach-arXiv-2022}
Robin Rombach, Andreas Blattmann, and Bj{\"o}rn Ommer.
\newblock {Text-Guided Synthesis of Artistic Images with Retrieval-Augmented
  Diffusion Models}.
\newblock \emph{arXiv preprint arXiv:2207.13038}, 2022{\natexlab{b}}.

\bibitem[Saharia et~al.(2022{\natexlab{a}})Saharia, Chan, Chang, Lee, Ho,
  Salimans, Fleet, and Norouzi]{saharia-SIGGRAPH-2022}
Chitwan Saharia, William Chan, Huiwen Chang, Chris Lee, Jonathan Ho, Tim
  Salimans, David Fleet, and Mohammad Norouzi.
\newblock {Palette: Image-to-image diffusion models}.
\newblock In \emph{Proceedings of SIGGRAPH}, pages 1--10, 2022{\natexlab{a}}.

\bibitem[Saharia et~al.(2022{\natexlab{b}})Saharia, Chan, Saxena, Li, Whang,
  Denton, Ghasemipour, Gontijo~Lopes, Karagol~Ayan, Salimans,
  et~al.]{Saharia-NeurIPS-2022}
Chitwan Saharia, William Chan, Saurabh Saxena, Lala Li, Jay Whang, Emily~L
  Denton, Kamyar Ghasemipour, Raphael Gontijo~Lopes, Burcu Karagol~Ayan, Tim
  Salimans, et~al.
\newblock Photorealistic text-to-image diffusion models with deep language
  understanding.
\newblock In \emph{Proceedings of NeurIPS}, pages 36479--36494,
  2022{\natexlab{b}}.

\bibitem[Saharia et~al.(2022{\natexlab{c}})Saharia, Ho, Chan, Salimans, Fleet,
  and Norouzi]{saharia-TPAMI-2022}
Chitwan Saharia, Jonathan Ho, William Chan, Tim Salimans, David~J. Fleet, and
  Mohammad Norouzi.
\newblock Image super-resolution via iterative refinement.
\newblock \emph{IEEE Transactions on Pattern Analysis and Machine
  Intelligence}, 45\penalty0 (4):\penalty0 4713--4726, 2022{\natexlab{c}}.

\bibitem[Salimans and Ho(2022)]{salimans-arXiv-2022}
Tim Salimans and Jonathan Ho.
\newblock Progressive distillation for fast sampling of diffusion models.
\newblock In \emph{Proceedings of ICLR}, 2022.

\bibitem[Schuhmann(2022)]{Schuhmann-laion-2022}
Christoph Schuhmann.
\newblock {LAION-Aesthetics}.
\newblock https://laion.ai/blog/laion-aesthetics/, 2022.

\bibitem[Schulman et~al.(2017)Schulman, Wolski, Dhariwal, Radford, and
  Klimov]{schulman-arXiv-2017}
John Schulman, Filip Wolski, Prafulla Dhariwal, Alec Radford, and Oleg Klimov.
\newblock Proximal policy optimization algorithms.
\newblock \emph{arXiv preprint arXiv:1707.06347}, 2017.

\bibitem[Schwartz et~al.(2020)Schwartz, Dodge, Smith, and
  Etzioni]{Schwartz-CACM-2020}
Roy Schwartz, Jesse Dodge, Noah~A. Smith, and Oren Etzioni.
\newblock {Green AI}.
\newblock \emph{Communications of the ACM}, 63\penalty0 (12):\penalty0
  54--–63, 2020.

\bibitem[Sinha et~al.(2020)Sinha, Garg, and Larochelle]{Sinha-NIPS-2020}
Samarth Sinha, Animesh Garg, and Hugo Larochelle.
\newblock {Curriculum by Smoothing}.
\newblock In \emph{Proceedings of NeurIPS}, pages 21653--21664, 2020.

\bibitem[Sohl-Dickstein et~al.(2015)Sohl-Dickstein, Weiss, Maheswaranathan, and
  Ganguli]{sohl-icml-2015}
Jascha Sohl-Dickstein, Eric Weiss, Niru Maheswaranathan, and Surya Ganguli.
\newblock Deep unsupervised learning using non-equilibrium thermodynamics.
\newblock In \emph{Proceedings of ICML}, pages 2256--2265, 2015.

\bibitem[Song et~al.(2021{\natexlab{a}})Song, Meng, and Ermon]{song-ICLR-2021b}
Jiaming Song, Chenlin Meng, and Stefano Ermon.
\newblock {Denoising Diffusion Implicit Models}.
\newblock In \emph{Proceedings of ICLR}, 2021{\natexlab{a}}.

\bibitem[Song and Dhariwal(2024)]{Song-ICLR-2024}
Yang Song and Prafulla Dhariwal.
\newblock {Improved Techniques for Training Consistency Models}.
\newblock In \emph{Proceedings of ICLR}, 2024.

\bibitem[Song and Ermon(2019)]{song-NeurIPS-2019}
Yang Song and Stefano Ermon.
\newblock Generative modeling by estimating gradients of the data distribution.
\newblock In \emph{Proceedings of NeurIPS}, pages 11918--11930, 2019.

\bibitem[Song et~al.(2021{\natexlab{b}})Song, Sohl-Dickstein, Kingma, Kumar,
  Ermon, and Poole]{song-ICLR-2021}
Yang Song, Jascha Sohl-Dickstein, Diederik~P. Kingma, Abhishek Kumar, Stefano
  Ermon, and Ben Poole.
\newblock {Score-Based Generative Modeling through Stochastic Differential
  Equations}.
\newblock In \emph{Proceedings of ICLR}, 2021{\natexlab{b}}.

\bibitem[Song et~al.(2023)Song, Dhariwal, Chen, and Sutskever]{Song-ICML-2023}
Yang Song, Prafulla Dhariwal, Mark Chen, and Ilya Sutskever.
\newblock Consistency models.
\newblock In \emph{Proceedings of ICML}, pages 32211--32252, 2023.

\bibitem[Soviany et~al.(2020)Soviany, Ardei, Ionescu, and
  Leordeanu]{soviany-wacv-2020}
Petru Soviany, Claudiu Ardei, Radu~Tudor Ionescu, and Marius Leordeanu.
\newblock {Image difficulty curriculum for generative adversarial networks
  (CuGAN)}.
\newblock In \emph{Proceedings of WACV}, pages 3463--3472, 2020.

\bibitem[Soviany et~al.(2021)Soviany, Ionescu, Rota, and
  Sebe]{Soviany-CVIU-2021}
Petru Soviany, Radu~Tudor Ionescu, Paolo Rota, and Nicu Sebe.
\newblock Curriculum self-paced learning for cross-domain object detection.
\newblock \emph{Computer Vision and Image Understanding}, 204:\penalty0
  103--166, 2021.

\bibitem[Soviany et~al.(2022)Soviany, Ionescu, Rota, and
  Sebe]{Soviany-IJCV-2022}
Petru Soviany, Radu~Tudor Ionescu, Paolo Rota, and Nicu Sebe.
\newblock Curriculum learning: A survey.
\newblock \emph{International Journal of Computer Vision}, 130\penalty0
  (6):\penalty0 1526--1565, 2022.

\bibitem[Strubell et~al.(2019)Strubell, Ganesh, and
  Mccallum]{Strubell-ACL-2019}
Emma Strubell, Ananya Ganesh, and Andrew Mccallum.
\newblock {Energy and Policy Considerations for Deep Learning in NLP}.
\newblock In \emph{Proceedings of ACL}, pages 3645--3650, 2019.

\bibitem[Vahdat et~al.(2021)Vahdat, Kreis, and Kautz]{vahdat-NeurIPS-2021}
Arash Vahdat, Karsten Kreis, and Jan Kautz.
\newblock Score-based generative modeling in latent space.
\newblock In \emph{Proceedings of NeurIPS}, pages 11287--11302, 2021.

\bibitem[Wallace et~al.(2024)Wallace, Dang, Rafailov, Zhou, Lou, Purushwalkam,
  Ermon, Xiong, Joty, and Naik]{Wallace-arxiv-2023}
Bram Wallace, Meihua Dang, Rafael Rafailov, Linqi Zhou, Aaron Lou, Senthil
  Purushwalkam, Stefano Ermon, Caiming Xiong, Shafiq Joty, and Nikhil Naik.
\newblock {Diffusion Model Alignment Using Direct Preference Optimization}.
\newblock In \emph{Proceedings of CVPR}, pages 8228--8238, 2024.

\bibitem[Wang et~al.(2023)Wang, Yue, Lu, Liu, Zhong, Song, and
  Huang]{Wang-ICCV-2023}
Yulin Wang, Yang Yue, Rui Lu, Tianjiao Liu, Zhao Zhong, Shiji Song, and Gao
  Huang.
\newblock {EfficientTrain: Exploring Generalized Curriculum Learning for
  Training Visual Backbones}.
\newblock In \emph{Proceedings of ICCV}, pages 5852--5864, 2023.

\bibitem[Wei et~al.(2021)Wei, Suriawinata, Ren, Liu, Lisovsky, Vaickus, Brown,
  Baker, Nasir-Moin, Tomita, Torresani, Wei, and Hassanpour]{Wei-WACV-2021}
Jerry Wei, Arief Suriawinata, Bing Ren, Xiaoying Liu, Mikhail Lisovsky, Louis
  Vaickus, Charles Brown, Michael Baker, Mustafa Nasir-Moin, Naofumi Tomita,
  Lorenzo Torresani, Jason Wei, and Saeed Hassanpour.
\newblock {Learn like a Pathologist: Curriculum Learning by Annotator Agreement
  for Histopathology Image Classification}.
\newblock In \emph{Proceedings of WACV}, pages 2472--2482, 2021.

\bibitem[Wu et~al.(2023)Wu, Hao, Sun, Chen, Zhu, Zhao, and Li]{Wu-arXiv-2023}
Xiaoshi Wu, Yiming Hao, Keqiang Sun, Yixiong Chen, Feng Zhu, Rui Zhao, and
  Hongsheng Li.
\newblock {Human Preference Score v2: A Solid Benchmark for Evaluating Human
  Preferences of Text-to-Image Synthesis}.
\newblock \emph{arXiv preprint arXiv:2306.09341}, 2023.

\bibitem[Wyatt et~al.(2022)Wyatt, Leach, Schmon, and
  Willcocks]{wyatt-CVPR-2022}
Julian Wyatt, Adam Leach, Sebastian~M. Schmon, and Chris~G. Willcocks.
\newblock {AnoDDPM: Anomaly Detection with Denoising Diffusion Probabilistic
  Models using Simplex Noise}.
\newblock In \emph{Proceedings of CVPR}, pages 650--656, 2022.

\bibitem[Xu et~al.(2024)Xu, Mi, Wang, and Chen]{xu-arXiv-2024}
Tianshuo Xu, Peng Mi, Ruilin Wang, and Yingcong Chen.
\newblock {Towards Faster Training of Diffusion Models: An Inspiration of A
  Consistency Phenomenon}.
\newblock \emph{arXiv preprint arXiv:2404.07946}, 2024.

\bibitem[Zhang et~al.(2023)Zhang, Rao, and Agrawala]{Zhang-ICCV-2023}
Lvmin Zhang, Anyi Rao, and Maneesh Agrawala.
\newblock {Adding Conditional Control to Text-to-Image Diffusion Models}.
\newblock In \emph{Proceedings of ICCV}, pages 3836--3847, 2023.

\bibitem[Ziegler et~al.(2020)Ziegler, Stiennon, Wu, Brown, Radford, Amodei,
  Christiano, and Irving]{Ziegler-arXiv-2020}
Daniel~M. Ziegler, Nisan Stiennon, Jeffrey Wu, Tom~B. Brown, Alec Radford,
  Dario Amodei, Paul Christiano, and Geoffrey Irving.
\newblock {Fine-Tuning Language Models from Human Preferences}.
\newblock \emph{arXiv preprint arXiv:1909.08593}, 2020.

\end{thebibliography}
}

\clearpage
\setcounter{page}{1}
\maketitlesupplementary

%\section{Supplemental Material}

\section{Detailed Preliminaries}
\label{supp_prelim}

\textbf{Diffusion models.}
Diffusion models \cite{Croitoru-TPAMI-2023, ho-NeurIPS-2020, karras-CVPR-2020,kingma-NeurIPS-2021, rombach-CVPR-2022, sohl-icml-2015,  song-ICLR-2021, song-NeurIPS-2019,  vahdat-NeurIPS-2021} are a class of generative models trained to reverse a process that progressively inserts Gaussian noise across $T$ steps into the original data samples, transforming them into standard Gaussian noise. Formally, the forward process defined by:
\begin{equation}
\label{eq_forward_process}
        x_t = \alpha_t x_0 + \sigma_t \epsilon, \epsilon \sim \mathcal{N}(0, \mathbf{I})
\end{equation}
transforms the original samples $x_0 \sim p(x_0)$ into noisy versions $x_t$, following the noise schedule implied by the time-dependent predefined functions $(\alpha_t)_{t=1}^{T}$ and $(\sigma_t)_{t=1}^{T}$. The model is trained to estimate the noise $\epsilon$ added in the forward step defined in Eq.~\eqref{eq_forward_process}, by minimizing the following objective:
\begin{equation}
    \mathcal{L}_{simple} = \mathbb{E}_{t \sim \mathcal{U}(1, T), \epsilon \sim \mathcal{N}(0, \mathbf{I}), x_0 \sim p(x_0)} \!\left\lVert \epsilon_t-\epsilon_\theta(x_t,t)\right\rVert^{2}.
\end{equation}
The generation process involves denoising, starting from a sample of standard Gaussian noise, denoted as \( x_T \sim \mathcal{N}(0, \mathbf{I}) \). It then follows the transitions outlined in Eq.~\eqref{eq_reverse_process} to produce novel samples:
\begin{equation}
    \label{eq_reverse_process}
    p_\theta(x_{t-1}|x_t) = \mathcal{N}\left(x_{t-1}; \mu_\theta(x_t, t), \sigma_{t|t-1}^2\frac{\sigma_{t-1}^2}{\sigma_t^2}\right), 
\end{equation}
with $\mu_\theta(x_t, t) = \frac{1}{\alpha_{t|t-1}}\left(x_t - \frac{\epsilon_\theta(x_t, t)\sigma_{t|t-1}^2}{\sigma_t} \right)$,
where $\sigma_{t|t-1}^2=\sigma_t^2-\alpha_{t|t-1}^2 \sigma_{t-1}^2$ and $\alpha_{t|t-1}=\frac{\alpha_t}{\alpha_{t-1}}$.

The forward process can also be defined in a continuous time manner \cite{song-ICLR-2021}, as a stochastic differential equation (SDE):
\begin{equation}
    \label{eq_forward_process_sde}
    \mathrm{d}x_t = f(t)x_t\mathrm{d}t + g(t)\mathrm{d}\omega_t, t \in [0, T],
\end{equation}
where, given the notations from Eq.~\eqref{eq_forward_process}, we can write $f(t) = \frac{\mathrm{d} \log{\alpha(t)}}{\mathrm{d}t}$ and $g^2(t) = \frac{\mathrm{d} \sigma^2(t)}{\mathrm{d}t} - 2 \frac{\mathrm{d} \log{\alpha(t)}}{\mathrm{d}t} \cdot \sigma^2(t)$, and $\omega_t$ is the standard Brownian motion.

Furthermore, the diffusion process described by the SDE from Eq.~\eqref{eq_forward_process_sde} can be reversed by another diffusion process given by a reverse-time SDE \cite{Andreson-SPA-1982, song-ICLR-2021}. In addition to this, \citet{song-ICLR-2021} showed that the reverse SDE has a corresponding ordinary differential equation (ODE), called Probability flow-ODE (PF-ODE), with the following form:
\begin{equation}
    \label{eq_ODE_reverse}
    \mathrm{d}x_t = f(t)x_t\mathrm{d}t + \frac{g^2(t)}{2 \sigma(t)}\epsilon_{\theta}(x_t, t).
\end{equation}

\begin{algorithm*}[!t]
\caption{Curriculum DPO (for diffusion models)}
\label{alg:method_diffusion}
%\begin{algorithmic}
%\small{
\KwIn{$\{(x_{0, i}, c)\}_{i=1}^M$ - the training samples, $r_\varphi(x_0, c)$ - the reward model which can be conditioned on $c$, $B$ - the number of batches for splitting the set of pairs, $\alpha_t, \sigma_t$ - the parameters of the noise schedule, $T$ - the last time step of diffusion, $\beta$ - DPO hyperparameter to control the divergence from the initial pre-trained state, $\sigma$ - the sigmoid function, $\eta$ - the learning rate, $\{H_k\}_{k=1}^B$ - the number of training iterations after including the $k$-th batch.}
\KwOut{
$\theta$ - the trained weights of the generative model.
}
$\hat{X} \leftarrow \{(x_{0,i}, c)|r_\varphi(x_{0,i}, c) \leq r_\varphi(x_{0,i-1},  c), i=\{2,3,...,M\}\};$ \textcolor{commentGreen}{$\lhd$ sort the samples in descending order of the rewards}\\
$S \leftarrow \left\{(x_{0,i}, x_{0, j}, c)| i,j \in \{1, \dots M\}; i<j; x_{0,i}, x_{0, j} \in \hat{X},  r_\varphi(x_{0,i}, c) > r_\varphi(x_{0,j}^l, c)  \right\}$; \textcolor{commentGreen}{$\lhd$ create pairs of examples using the order from $\hat{X}$}\\
$\mathrm{L}_k \leftarrow \left\{\frac{(M-1) \cdot(B-k)}{B}\right\}_{k=1}^B$; \textcolor{commentGreen}{$\lhd$ the minimum preference limits of the batches}\\
$\mathrm{R}_k \leftarrow \left\{\frac{(M-1)\cdot(B-(k-1))}{B}\right\}_{k=1}^B$;  \textcolor{commentGreen}{$\lhd$ the maximum preference limits of the batches}\\
$S_k\leftarrow \left\{(x_{0}^w, x_{0}^l, c) | (x_{0}^w, x_{0}^l) = (x_{0,i}, x_{0, j}); \mathrm{L}_k   < j-i \leq \mathrm{R}_k  ; (x_{0,i}, x_{0, j}, c) \in S \right\}_{k = 1}^B$;  $\lhd$  \textcolor{commentGreen}{the batches of increasingly difficult pairs}\\
$P \leftarrow \emptyset$;  \textcolor{commentGreen}{$\lhd$ current training set} \\
\ForEach{$k \in \{1, \dots, B\}$}
{
    $P \leftarrow P \cup S_k$; \textcolor{commentGreen}{$\lhd$ include a new batch in the training}

    \ForEach{$i \in \{1, \dots, H_k\}$}
    {
    $(x_0^w, x_0^l, c) \sim \mathcal{U}(P)$; 
    $t \sim \mathcal{U}\{1, \dots, T\}$; $\epsilon^w, \epsilon^l \sim \mathcal{N}(0, \mathbf{I})$;\\
    $x_{t}^w \leftarrow \alpha_{t} x_0^w + \sigma_{t} \epsilon^w$;  \textcolor{commentGreen}{$\lhd$ forward process} \\
    $x_{t}^l \leftarrow \alpha_{t} x_0^l + \sigma_{t} \epsilon^l$; \textcolor{commentGreen}{$\lhd$ forward process}\\
    $\mathcal{L}_{\mathrm{\mbox{\scriptsize{Diff-DPO}}}}(\theta) \leftarrow - \Big[\log\sigma\Big(\!-\beta T %\omega(\lambda_t)
        \Big(
    \left(\lVert\epsilon^w - \rVert_2^2 -
    \lVert\epsilon^w - \epsilon_{\mbox{\scriptsize{ref}}}^w(x_t^w, t, c)\rVert_2^2
    \right) -  
    \left(\lVert\epsilon^l - \epsilon_\theta^l(x_t^l, t, c)\rVert_2^2 -
    \lVert\epsilon^l - \epsilon_{\mbox{\scriptsize{ref}}}^l(x_t^l, t, c)\rVert_2^2
    \right)
    \Big) \Big) \Big]$; \textcolor{commentGreen}{$\lhd$ DPO loss}\\
    $\theta \leftarrow \theta - \eta \frac{\partial \mathcal{L}_\mathrm{\mbox{\tiny{Diff-DPO}}}}{\partial \theta}$; \textcolor{commentGreen}{$\lhd$ update the weights}
    }   
}
\textbf{return} $\theta$
%}
%\end{algorithmic}
\end{algorithm*}

\noindent
\textbf{Consistency models.} Consistency models \cite{Luo-arXiv-2023, Song-ICLR-2024,  Song-ICML-2023} are a new class of generative models. These models operate on the idea of training a model to associate each point along a trajectory of the PF-ODE (Eq.~\eqref{eq_ODE_reverse}) to the trajectory's initial point, which corresponds to the denoised sample. Such models can either be trained from scratch or through distillation from a pre-trained diffusion model. In our study, we employ the distillation method, so we next detail this approach.

Given a solution trajectory $\{x_t\}_{t\in \left[\delta, T\right]}$ of the PF-ODE defined in Eq.~\eqref{eq_ODE_reverse}, where $\delta \rightarrow 0$, the training of a consistency model $f_\phi(x_t, t)$ involves enforcing the self-consistency property across this trajectory, such that, $\forall t, t' \in \left[ \delta, T \right]$, the condition $f_\phi(x_t, t) = f_\phi(x_{t'}, t')$ holds. The loss function designed to achieve this self-consistency is described as follows: 
\begin{equation}
\label{app_eq_consistency_distillation}
    \mathcal{L}_{\mbox{\scriptsize{CD}}}(\phi) = d(f_\phi(x_{t_{n+1}}, t_{n+1}), f_{\phi^{-}}(\hat{x}_{t_n}^{\theta}, t_n)),
\end{equation}
where $d$ is a distance metric, $n \sim \mathcal{U}(1, N)$, $N$ is the discretization length of the interval $\left[0, T\right]$, $\phi$ are the trainable parameters of the consistency model and $\phi^{-}$ is a running average of $\phi$. The term $\hat{x}_{t_n}^\theta$ represents a one-step denoised version of $x_{t_{n+1}}$, obtained by applying an ODE solver on the PF-ODE. The solver operates using a pre-trained diffusion model, $\epsilon_\theta(x_{t_n}, t_n)$.

\noindent
\textbf{Direct Preference Optimization (DPO).}
Training pipelines based on Reinforcement Learning with Human Feedback (RLHF) \cite{Ziegler-arXiv-2020} have been highly successful in aligning Large Language Models to human preferences. These pipelines feature an initial phase where a reward model is trained using examples ranked by humans, followed by a reinforcement learning phase where the policy model is fine-tuned to align with the learned reward model. In this context, \citet{Rafailov-NeurIPS-2023} introduced DPO as an alternative to the previous pipeline, which bypasses the training of the reward model and directly optimizes the policy model using the ranked examples.

The training dataset contains triplets of the form $(c, x_0^w, x_0^l)$, where $x_0^w$ denotes the favored sample, $x_0^l$ the unfavored one and $c$ is a condition used to generate both samples. RLHF trains a reward model by maximizing the likelihood $p(x_0^w \prec x_0^l|c)$\footnote{$a \prec b$ denotes that $a$ precedes $b$ in the ranking implied by the reward model.}, which, under the Bradley-Terry (BT) model, has the following form:
\begin{equation}
    p_{\mbox{\scriptsize{BT}}}(x_0^w \prec x_0^l|c) = \sigma(r_\varphi(x_0^w, c) - r_\varphi(x_0^l, c)),
\end{equation}
where $\sigma$ denotes the sigmoid function and $r_\varphi$ is the reward model parameterized by the trainable parameters $\varphi$. The training objective for the reward model is the negative log-likelihood:
\begin{equation}
    \label{eq_neg_log}
    \mathcal{L}_{\mbox{\scriptsize{BT}}} = - \mathbb{E}_{x_0^w, x_0^l, c}\left[\log{\sigma(r_\varphi(x_0^w, c) - r_\varphi(x_0^l, c))}\right].
\end{equation}
After training the reward model $r_\varphi(x_0, c)$, RLHF optimizes a conditional generative model $p_\theta(x_0|c)$ to maximize the reward $r_\varphi(x_0, c)$ and, at the same time, controls the deviance from a reference model $p_{\mbox{\scriptsize{ref}}}(x_0, c)$ through a Kullback–Leibler (KL) divergence term:
\begin{equation}
    \label{eq_policy_opt}
    \max_{\theta} \mathbb{E}_{c, x_0 \sim p_\theta(x_0|c)}\left[ r_\varphi(x_0, c)\!-\!\beta \mbox{KL}(p_\theta(x_0|c), p_{\mbox{\scriptsize{ref}}}(x_0| c))\right]\!,
\end{equation}
where $\beta$ controls the importance of the divergence term.

To derive the DPO objective, \citet{Rafailov-NeurIPS-2023} write the optimal policy model $p_\theta^*$ of Eq.~\eqref{eq_policy_opt} as a function of the reward and reference model, as shown in prior works \cite{Peng-arXiv-2019,Peters-ICML-2007}:
\begin{equation}
    \label{eq_optimal_policy}
    p_{\theta}^{*}(x_0|c) = \frac{p_{\mbox{\scriptsize{ref}}}(x_0|c) \cdot \exp\left(\frac{r(x_0, c)}{\beta}\right)}{Z(c)},
\end{equation}
where $Z(c) = \sum_{x_0}{p_{\mbox{\scriptsize{ref}}}(x_0|c) \cdot \exp\left(\frac{r(x_0, c)}{\beta}\right)}$ is a normalization constant. Further, from Eq.~\eqref{eq_optimal_policy}, \citet{Rafailov-NeurIPS-2023} rewrite the reward as:
\begin{equation}
    \label{eq_reward_function}
    r(x_0, c) = \beta\left( \log \frac{p_{\theta}^{*}(x_0|c)}{p_{\mbox{\scriptsize{ref}}}(x_0|c)} + \log Z(c)\right).
\end{equation}
Finally, the DPO objective is obtained after replacing the reward in Eq.~\eqref{eq_neg_log} with the form from Eq.~\eqref{eq_reward_function}:
\begin{equation}
    \label{app_eq_dpo}
        \begin{split}
        \mathcal{L}_\mathrm{\mbox{\scriptsize{DPO}}}(\theta)\!\!=\!\!-\!\mathbb{E}_{x_0^w\!, x_0^l, c} \!\left[\!\log\!\sigma\!\bigg(\!\!\beta\!\! %\cdot
        \left(\!\!\log\!{\frac{p_\theta(x_0^w\!|c)}{p_{\mbox{\scriptsize{ref}}}(x_0^w\!|c)}}\!-\!\log\!{\frac{p_\theta(x_0^l\!|c)}{p_{\mbox{\scriptsize{ref}}}(x_0^l\!|c)}}\!\!\right)\!\!\!\!\bigg)\!\!\right]\!\!,
    \end{split} 
\end{equation}
To grasp the intuition behind $\mathcal{L}_\mathrm{\mbox{\scriptsize{DPO}}}$, we can analyze  its gradient with respect to $\theta$:
\begin{equation}
    \label{app_eq_grad_dpo}
    \begin{split}
            \frac{\partial \mathcal{L}_\mathrm{\mbox{\scriptsize{DPO}}}(\theta)}{\partial\theta}\!=\!&-\beta \mathbb{E}_{x_0^w, x_0^l, c}\Bigg[\sigma\left( \hat{r}_\theta(x_0^l, c) - \hat{r}_\theta(x_0^w, c)\right)\!\cdot\!\\&\left( \frac{\partial\log{p_\theta(x_0^w|c)}}{\partial\theta} - \frac{\partial\log{p_\theta(x_0^l|c)}}{\partial\theta} \right)\!\Bigg],
    \end{split}
\end{equation}
with $\hat{r}_{\theta}(x_0,c)=\beta\cdot\log{\frac{p_\theta(x_0|c)}{p_{\mbox{\scriptsize{ref}}}(x_0|c)}}$. By analyzing Eq.~\eqref{app_eq_grad_dpo}, as discussed in \cite{Rafailov-NeurIPS-2023}, it is evident that the DPO objective enhances the likelihood of favored examples, while diminishing it for the unfavored ones. The magnitude of the update is proportional to the error in $\hat{r}_\theta$. Here, the term ``error'' refers to the degree to which $\hat{r}_\theta$ incorrectly prioritizes the sample $x_0^l$.

\begin{figure*}
  \centering
  \includegraphics[width=1.\linewidth]{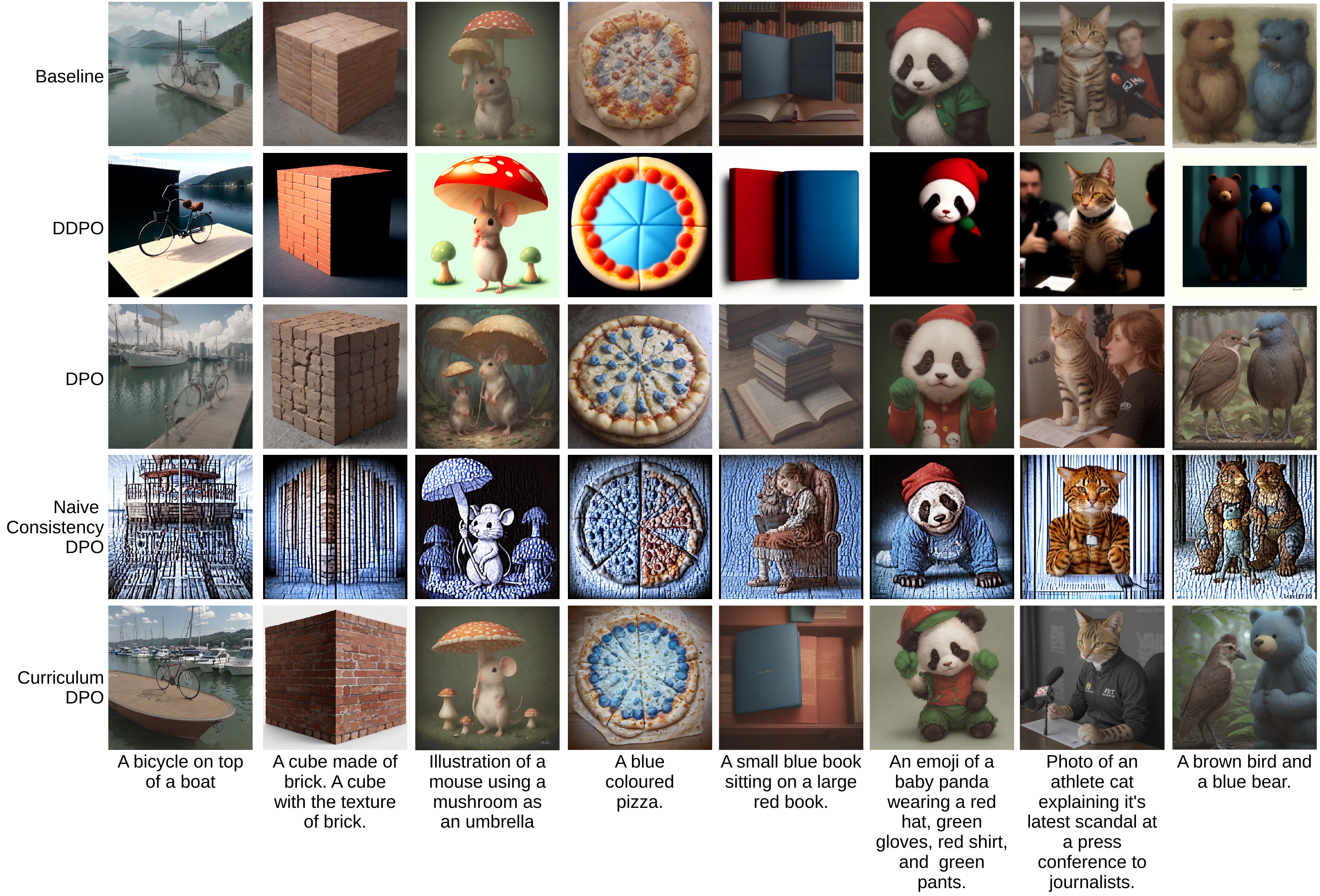}
  \vspace{-0.7cm}
   \caption{Qualitative results before and after fine-tuning for the text alignment task on DrawBench. The fine-tuning methods are: DDPO, DPO, Naive DPO and Curriculum DPO. Best viewed in color.}
   \vspace{-0.3cm}
   \label{drawbnech_qualitative_text_align}
\end{figure*}

\section{Curriculum DPO for Diffusion Models}
\label{supp_alg}
We formally present the application of Curriculum DPO to diffusion models in Algorithm~\ref{alg:method_diffusion}. The initial steps 1-9, which outline the curriculum strategy, are identical with those used in the implementation for consistency models described in Algorithm~\ref{alg:method}. Steps 10-14 are changed to include the forward process for the preferred and less preferred samples, along with the Diffusion-DPO loss defined in Eq.~\eqref{eq_diffusion_dpo}.

\begin{figure*}[t]  
  \centering
  \includegraphics[width=1.\linewidth]{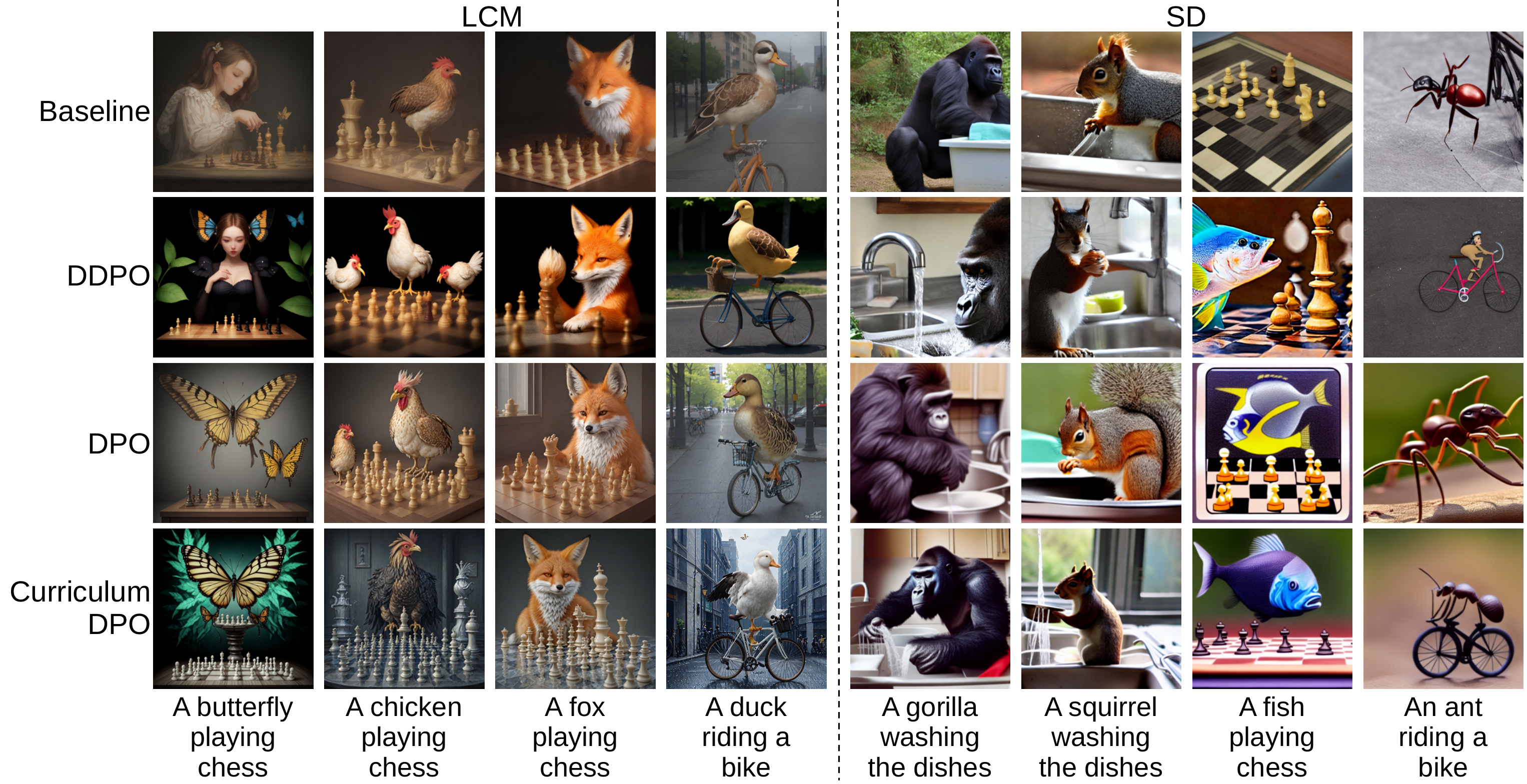}
  \vspace{-0.6cm}
  \caption{Qualitative results after fine-tuning with HPSv2 as the reward model (human preference). The fine-tuning alternatives are: DDPO, DPO and Curriculum DPO. Best viewed in color.}
  \label{qualitative_human_preference}
\end{figure*}

\section{Importance of Consistency-DPO}

Our work makes two contributions: Curriculum DPO and Consistency-DPO. While the novelty and importance of Curriculum DPO is more obvious, we consider that the significance of Consistency-DPO is not immediately observable. To this end, it is important to note that the Diffusion-DPO \cite{Wallace-arxiv-2023} approach cannot be directly applied to consistency models. The most direct modification is to substitute the noise estimation in the Diffusion-DPO objective with the consistency distillation loss used in consistency models. However, applying this modification directly breaks the consistency property required by these models and leads to poor results, as shown in Figure~\ref{drawbnech_qualitative_text_align} and further discussed in Section \ref{supp_qual}.

We found two solutions for this problem. The first is to reintegrate the consistency distillation for both preferred and non-preferred samples as separate components within the optimization function, in addition to the Consistency-DPO loss (Eq.~\eqref{eq_consistency_dpo}). This method, however, introduces the need for additional hyperparameters to balance these terms, which represents a significant drawback because it requires extensive hyperparameter tuning.

The second solution, which we ultimately adopt in our study, is to ensure the initial estimation for the ODE's starting point (the target in the consistency distillation loss) is a sample of the consistency model that undergoes fine-tuning. We accomplish this by replacing the Exponential Moving Average (EMA) model, that is typically used to get this estimation, with the pre-trained model from which we begin the fine-tuning process. This approach maintains the integrity of the consistency model's properties throughout the training. 

We thus conclude that adapting DPO to consistency models is not trivial, since the adaptation requires a deep understanding of the framework and strong knowledge about the use of gradients. 

\section{More Quantitative Results}
\label{supp_quant}

\begin{table}[!t]

  \centering
  \setlength\tabcolsep{0.44em}
  \small{
  \begin{tabular}{cccc}
  \toprule
    %       & \multicolumn{3}{c}{Dataset Pick-a-Pic}   \\ 
    % \cmidrule(l{2pt}r{2pt}){2-4}
     {Fine-Tuning}     & Text  & \multirow{2}{*}{Aesthetics} & Human    \\  
       Strategy   &  Alignment &  & Preference   \\
    \midrule
   -  & 0.5246 & 5.6675 & 0.2673   \\
   DDPO & 0.5317 &
  5.6764 &
  0.2717 
   \\
    
   DPO & 0.5328 & 5.7593 & 0.2725  \\

     Curriculum DPO (ours) & \textbf{0.5413} & \textbf{5.7998} & \textbf{0.2783}  \\

  \bottomrule
  \end{tabular}
  }\vspace{-0.2cm}
    \caption{Text alignment, aesthetic and human preference scores on Pick-a-Pic ($D_3$), obtained by the baseline (pre-trained) Stable Diffusion model versus the three fine-tuning strategies: DDPO, DPO and Curriculum DPO. The best scores are highlighted in bold.}
  \label{tab_exp_pickapic}
  %\vspace{-0.4cm}
\end{table}

\noindent
\textbf{Results on Pick-a-Pic.}
We report additional results for Stable Diffusion on 150,000 image pairs from Pick-a-Pic ($D_3$) in Table \ref{tab_exp_pickapic}. In this scenario, the dataset already includes pairs of winning and losing images, so we only apply the reward models for the ranking described in Figure~\ref{fig_pipeline}. The results reported on $D_3$ are consistent with those reported on  ${D}_1$ and ${D}_2$, further highlighting the importance of curriculum learning.

\begin{table}[!t]
  \centering
  \setlength\tabcolsep{0.46em}
  \small{
  \begin{tabular}{ccc}
  \toprule
    %       & \multicolumn{2}{c}{Reward Model}   \\ 
    % \cmidrule(l{2pt}r{2pt}){2-3}
     {Fine-Tuning} Strategy    & {LLaVA}  & {Phi-3} \\  
    \midrule
   -  & 0.6804 & 0.6804    \\
   DDPO & 0.7629 & 0.7602 
   \\
    
DPO & 0.7614 & 0.7643   \\
Curriculum DPO (ours) & \textbf{0.7703} & \textbf{0.7736}  \\

  \bottomrule
  \end{tabular}
  }\vspace{-0.2cm}
    \caption{Text alignment results on dataset $D_1$ by using two reward models (LLaVA and Phi-3) for DDPO, DPO and Curriculum DPO applied on Stable Diffusion. The best scores are highlighted in bold.}
  \label{tab_exp_phi3}
  \vspace{-0.4cm}
\end{table}

\noindent
\textbf{Results with different reward models.}
In Table~\ref{tab_exp_phi3}, we compare text alignment results for two alternative reward models: LLaVA \cite{liu-NeurIPS-2024} and Phi-3 \cite{abdin-arxiv-2024}. During training, we use a reward model to extract image descriptions and then measure their similarity to the original prompts to produce winning and losing image pairs. The same similarity scores also determine the ranking used by Curriculum DPO. These experiments further confirm the superiority of Curriculum DPO over DPO and DDPO, regardless of the employed reward model. 

\section{More Qualitative Results}
\label{supp_qual}

\begin{figure*}[t] 
  \centering
  \includegraphics[width=1.\linewidth]{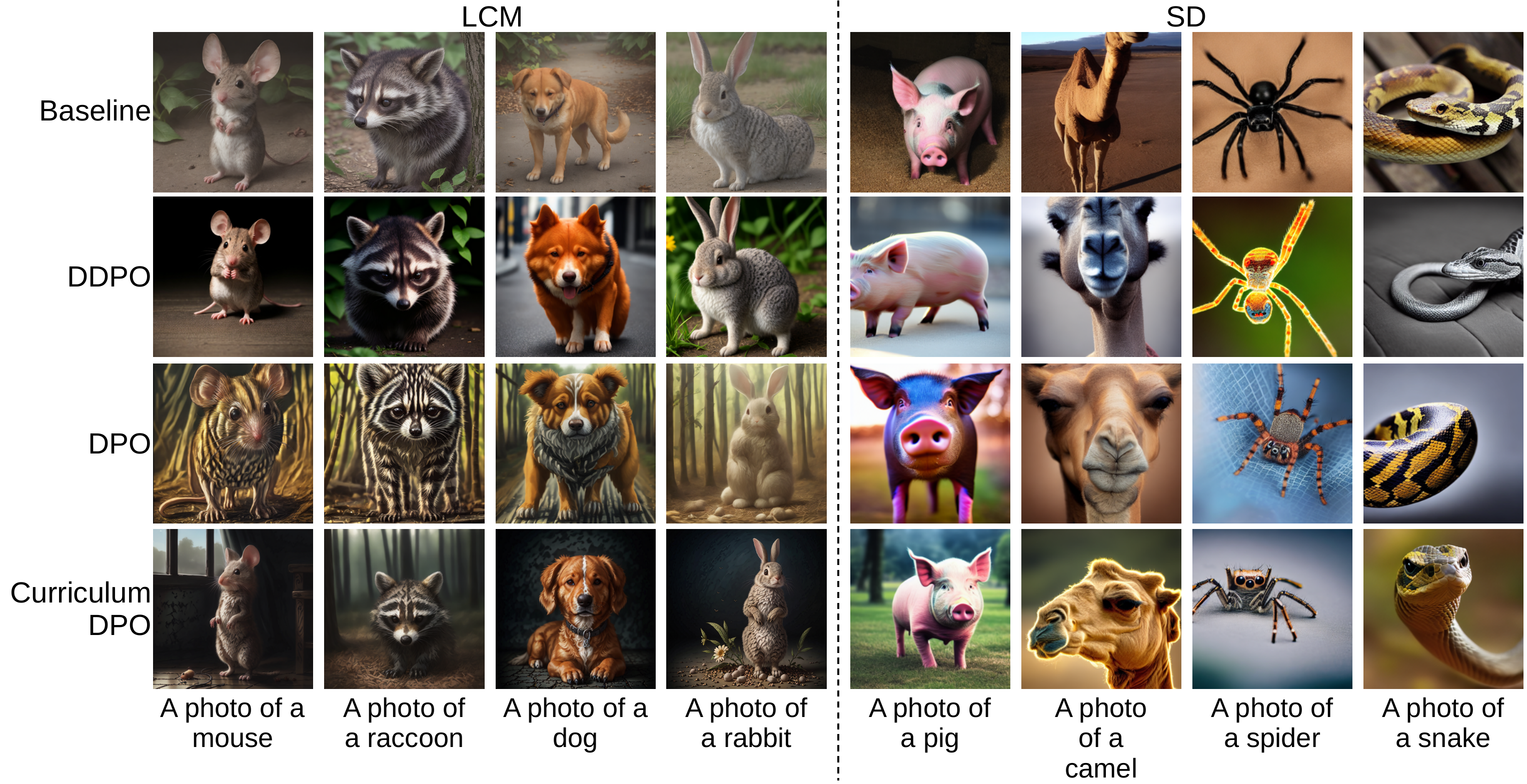}
  \vspace{-0.6cm}
  \caption{Qualitative results after fine-tuning with the LAION Aesthetics Predictor as the reward model. The fine-tuning alternatives are: DDPO, DPO and Curriculum DPO. Best viewed in color.}
  \label{qualitative_aesthetics}
\end{figure*}

In Figures~\ref{qualitative_human_preference} and \ref{qualitative_aesthetics}, we present qualitative results after fine-tuning the models with HPSv2 and LAION Aesthetics Predictor as reward models on $D_1$, respectively. Fine-tuning for human preference (Figure~\ref{qualitative_human_preference}) generally results in generating images with more details for the LCM model. Curriculum DPO, in particular, produces better aesthetics for both foreground objects and the background. In contrast, the SD results show a better alignment with the text prompt, Curriculum DPO being the only method that generates the ant on a bike displayed in the last column. Fine-tuning for improving the visual appeal (Figure \ref{qualitative_aesthetics}) returns in general, as expected, better aesthetics for the animals. However, Curriculum DPO returns several examples that look better, \eg~the camel in the sixth column and the dog in the third column.

\begin{figure*}[!t]
\begin{subfigure}[t]{.32\textwidth}
  \centering
  % include second image
  \includegraphics[width=.99\linewidth]{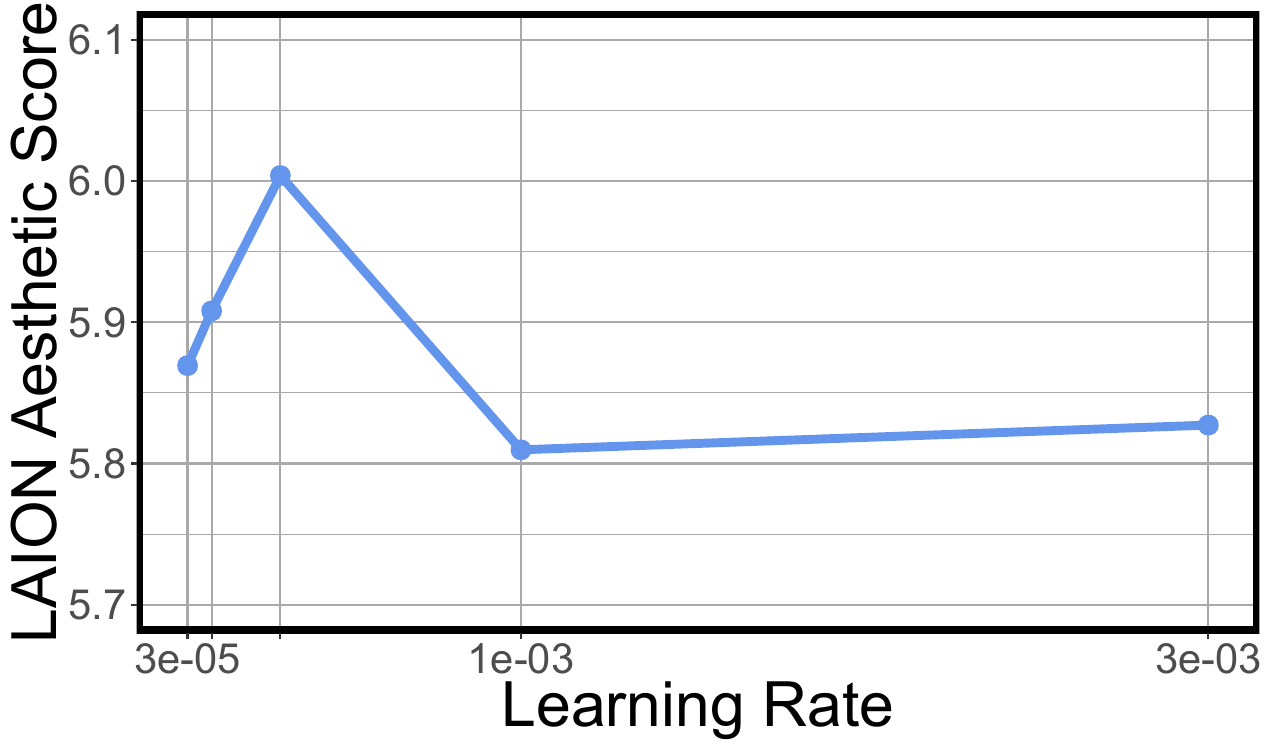}  
  \vspace{-0.4cm}
  \caption{Varying the learning rate for Curriculum DPO on LCM.}
  \label{fig:sub-lr}
\end{subfigure}
\hfill
\begin{subfigure}[t]{.32\textwidth}
  \centering
  % include third image
\includegraphics[width=.99\textwidth]{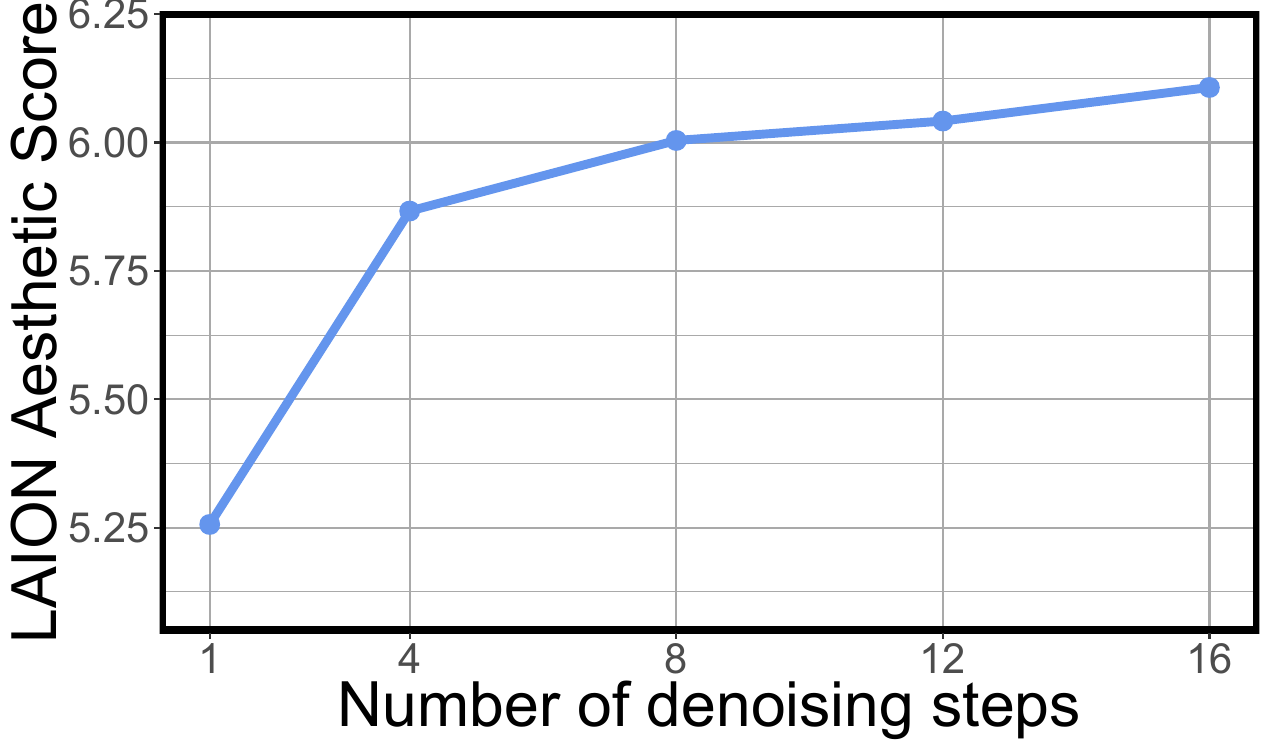}  
\vspace{-0.4cm}
  \caption{Varying the number of LCM generation steps for Curriculum DPO.}
  \label{fig:sub-denoising}
\end{subfigure}
\hfill
\begin{subfigure}[t]{.32\textwidth}
  \centering
  % include third image
\includegraphics[width=.99\textwidth]{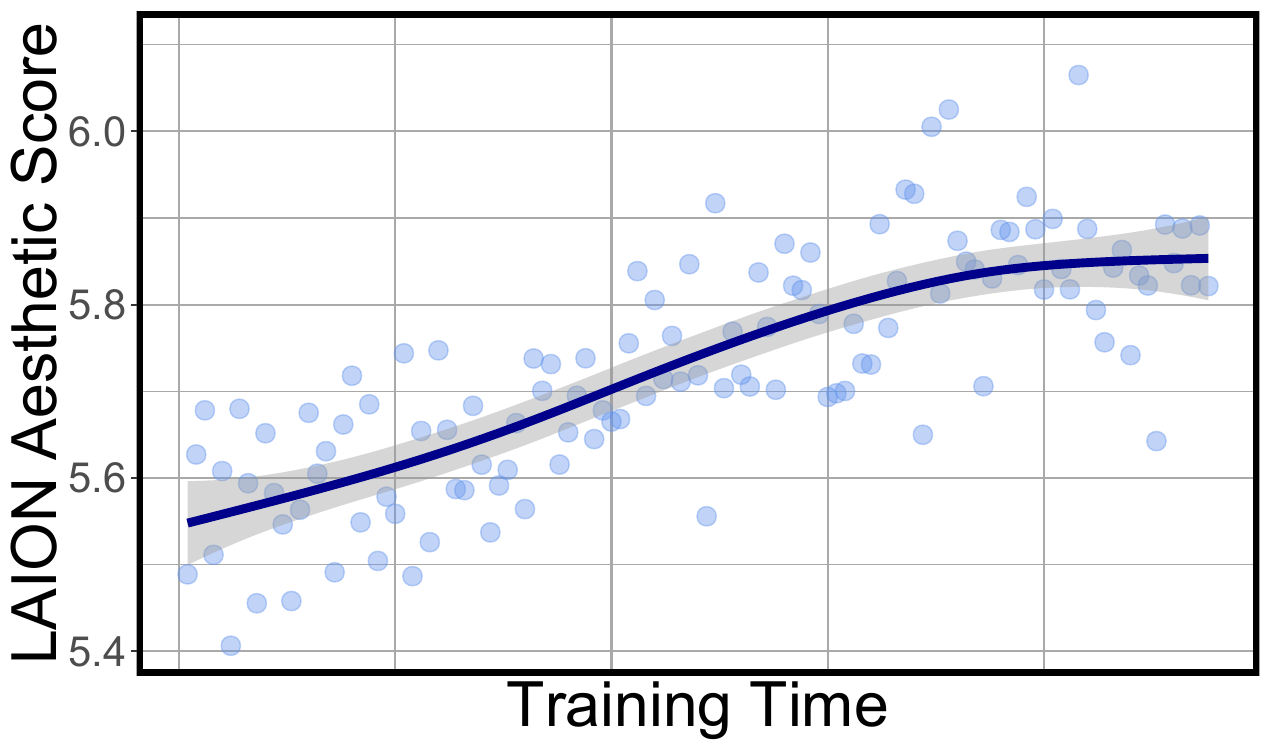}  
\vspace{-0.4cm}
  \caption{The progression of the aesthetic reward function during training with the DDPO method.}
  \label{fig:sub-aesthetic-reward}
\end{subfigure}
\vspace{-0.2cm}
\caption{Additional ablation results for Curriculum DPO applied on LCM are depicted in Figure~\ref{fig:sub-lr} and \ref{fig:sub-denoising}.  In Figure~\ref{fig:sub-aesthetic-reward}, we show the evolution of the reward score when training Stable Diffusion with DDPO. All experiments are carried out on DrawBench.}
%\vspace{-0.2cm}
\label{fig_rebuttal_ablation}
\end{figure*}

In Figure~\ref{drawbnech_qualitative_text_align}, we show qualitative results when fine-tuning the model for text alignment on $D_2$. In addition to the baseline, DPO, DDPO and Curriculum DPO results, we also include the images generated by a naive implementation of Consistency-DPO. This implementation refers to the most direct adaptation of Diffusion-DPO to consistency models. More precisely, we substitute the noise estimation in the Diffusion-DPO objective with the consistency distillation loss used in consistency models. However, applying this modification directly breaks the consistency property required by these models and leads to bad results, as illustrated in the 4th row of Figure~\ref{drawbnech_qualitative_text_align}.

\begin{figure*}[t]
  \centering
  \includegraphics[width=0.6\textwidth]{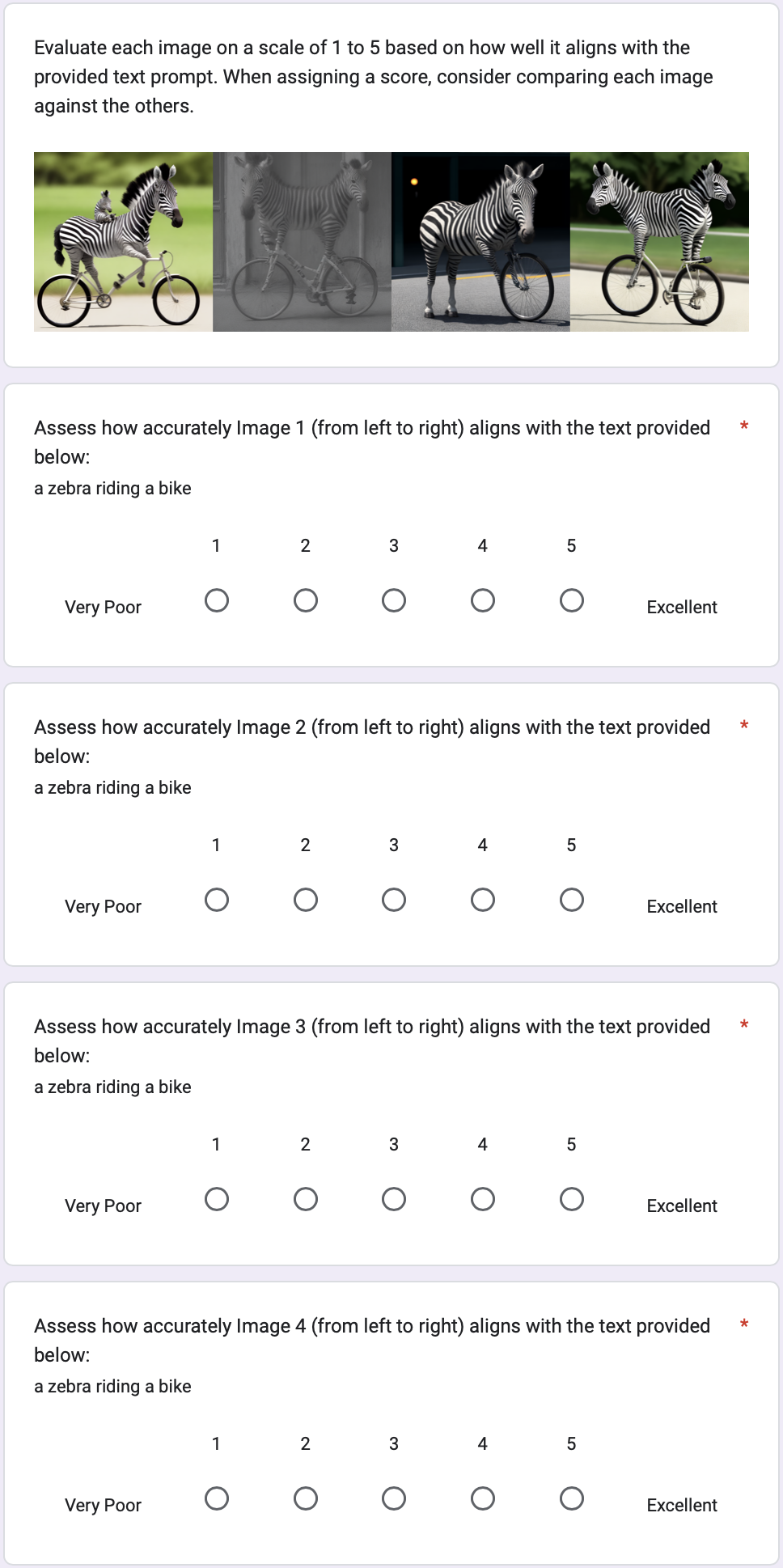}
  %\fbox{\rule[-.5cm]{0cm}{4cm} \rule[-.5cm]{4cm}{0cm}}
  \vspace{-0.1cm}
  \caption{A screenshot of one of the annotation forms, showcasing the annotation interface for one text prompt and the four corresponding images that are generated with alternative methods. The instructions are followed by the generated images, which are displayed on the same row, side by side. The images are placed in a random order to obfuscate the methods used to generate the images. A radio button list allows the users to input the rating for each generated image. Best viewed in color.\vspace{-0.5cm}}
  \label{fig_annotation_form}
\end{figure*}

\begin{table*}[t!]
  \centering
  \begin{tabular}{ccccc}
  \toprule
    Model     & Fine-Tuning Strategy     & Text Alignment & Aesthetics & Human Preference \\ 
    \midrule
    \multirow{4}{*}{LCM} & - & 0.5602 &  5.8038& 0.2610 \\
    &  DDPO  & 0.5627  & 5.9488  & 0.2780 \\

    &  DPO & 0.5639 & 5.9611 & 0.2783\\

    & Curriculum DPO (ours) & \textbf{0.5654} & \textbf{6.0038} & \textbf{0.2793} \\
  \bottomrule
  \end{tabular}
   \caption{Text alignment, aesthetic and human preference scores obtained on the DrawBench dataset by the baseline (pre-trained) LCM versus the three fine-tuning strategies: DDPO, DPO and Curriculum DPO. The DDPO, DPO and Curriculum DPO methods use only 5 images per prompt during optimization. The best scores are highlighted in bold.}
   \label{tab_five_images}
  \vspace{-0.3cm}
\end{table*}

\section{Additional Ablations}

Aside from the ablation results presented in Figure \ref{fig_ablation} and Table \ref{tab_ablation} from the main article, there are a few other hyperparameters involved in the training process, such as the learning rate and the number of steps used in the multi-step generation of LCM. We performed additional ablation studies on the learning rate and the number of steps used by LCM, on the DrawBench dataset, using $M=5$ generated images per prompt. The results presented in Figures~\ref{fig:sub-lr} and~\ref{fig:sub-denoising} demonstrate that, regardless of the chosen values, the outcomes consistently surpass the baseline (see Table \ref{tab_five_images}). We emphasize that we did not try to tune these hyperparameters for Curriculum DPO to avoid overfitting in hyperparameter space. Moreover, we underline that some apparent hyperparameters directly depend on already ablated hyperparameters. For example, the hyperparameter $B$ (ablated in Figure \ref{fig:sub-B}) is the only one that influences the minimum/maximum preference limits $\mathrm{L}_k$ and $\mathrm{R}_k$, which are computed in steps 3 and 4 of both Algorithm \ref{alg:method} and Algorithm \ref{alg:method_diffusion}. Note that the equations for $\mathrm{L}_k$ and $\mathrm{R}_k$ generate equally-sized batches, so the limits change only when we change the number of curriculum batches $B$. Therefore, ablating $\mathrm{L}_k$ and $\mathrm{R}_k$ is redundant.

% We configured the training parameters of DDPO to achieve a total of 10,000 network updates, similar to the number of updates we had for DPO and Curriculum DPO. For evaluation, we used the final checkpoint. 

In Figure~\ref{fig:sub-aesthetic-reward}, we present the evolution of the reward score during the Stable Diffusion training with DDPO. For DPO and Curriculum DPO, we did not preserve the reward curves, as these methods did not involve multiple queries to the reward models. Instead, they rely solely on the original example ranking throughout the entire training process. Thus, additional queries to the reward models are unnecessary for DPO and Curriculum DPO. This represents an advantage of these methods over DDPO. 

\section{Human Evaluation Study}
\label{supp_human}

In the human evaluation study, participants were asked to rate generated images from two perspectives: prompt alignment and aesthetics. The images were generated either with SD or LCM. We created a separate annotation form containing $80$ text prompts for each (task, generative model) pair, resulting in four independent annotation forms. For each generative architecture, there are four images per prompt: one from each fine-tuning strategy (DDPO, DPO and Curriculum DPO), along with another one corresponding to the pre-trained generative model. For each prompt, the images were displayed in a random order, preventing annotators from knowing which strategy was used to generate a certain image. The users were asked to rate each image with an integer grade between $1$ and $5$, as shown in Figure~\ref{fig_annotation_form}. The evaluation instructions were customized for each task. For text alignment, we requested the annotators to give their ratings based on how closely each generated image matches the accompanying text prompt. For aesthetics, the participants were asked to compare the images and rate each one according to their personal preference. 

Since there are four images for each prompt and an annotation form comprises 80 prompts, the number of images to be annotated in one form is 320. Each form was completed by nine human evaluators, yielding a total of 2,880 annotations per experiment (form). Since we conducted the study on two generative models (SD and LCM) and two tasks (prompt alignment and aesthetics), the total number of collected annotations is 11,520.

The average time to complete the annotations for a single form is around 15-20 minutes. The nine human annotators who agreed to complete the annotation forms are either close collaborators, family members or friends of the authors. They volunteered to perform the annotations for free. To make sure that the annotations are relevant, we computed the inter-rater agreement, obtaining a Kendall Tau correlation coefficient of $0.34$. This translates into $69.8\%$ of all image pairs being concordant among annotators. Additionally, we performed statistical testing for the evaluations, and found that the voting results are statistically significant, at a p-value below 0.005.

\section{Scalability}

In the ablation study presented in Figure~\ref{fig:sub-M} of the main paper, we examine the visual appeal reward when we vary the numbers of generated images per prompt. Here, we provide a more detailed analysis of the extreme case based on 5 images per prompt, comparing Curriculum DPO with all the other fine-tuning strategies across all the three studied tasks. The results shown in Table~\ref{tab_five_images} confirm that our method, Curriculum DPO, surpasses the competing methods even when the number of image samples per prompt is low. Therefore, we conclude that our training strategy does not require a high number of generated images per prompt to outperform DPO and DDPO.

\section{Limitations}
\label{sec: limitations}

One limitation of our model is the introduction of additional hyperparameters, such as $B$ or $K$. These might require tuning in order to find the optimal values, which involves more computing power. However, in the ablation study from Section~\ref{sec: experiments}, we demonstrate that Curriculum DPO outperforms all baselines for multiple hyperparameter combinations. Therefore, suboptimal hyperparameter choices can still improve the generative models.

A limitation of text-to-image generative models (as well as reward models) is the poor ability to disambiguate words in the input prompt. This can be observed especially in the prompt alignment task, where a word with multiple meanings or connotations leads to generating poor results. For example, the prompt ``a turkey riding a bike'' often results in images of a cooked meal instead of a live bird. Curriculum DPO does not address this generic limitation of generative and reward models.

\section{Broader Impact}
\label{sec: broader_impacts}

Generative models can be a valuable asset in many scenarios, ranging from boosting the productivity of creative tasks to being integrated in applications that are used on a daily basis (such as image restoration or super-resolution). Nevertheless, it might also represent a great source of fake data aiming for disinformation and impersonation, especially when the model is optimized to human preferences. In the recent years, an increase in deep fake materials flooded the Internet, with attackers aiming to spread false information or even steal sensitive information by posing as another entity or person.

While we strongly believe in the benefits of very capable generative models, we are aware of the potential risks. However, we can see that governments are working very closely with academia and industry on safely developing artificial intelligence, and thus observe and support the increasing focus on models that detect AI-generated content to mitigate the aforementioned risks. Notably, the ultimate goal of the project that funded our research is to develop robust deepfake detectors.
\end{document}